\documentclass[10pt,twocolumn,letterpaper]{article}

\usepackage[pagenumbers]{cvpr} 
\usepackage[accsupp]{axessibility} 
\usepackage[dvipsnames]{xcolor}

\definecolor{cvprblue}{rgb}{0.21,0.49,0.74}
\usepackage[pagebackref,breaklinks,colorlinks,citecolor=cvprblue]{hyperref}
\usepackage{microtype}
\usepackage{graphicx}
\usepackage{subcaption}
\usepackage{scalerel,xparse}
\NewDocumentCommand\emojione{}{\scalerel*{\includegraphics{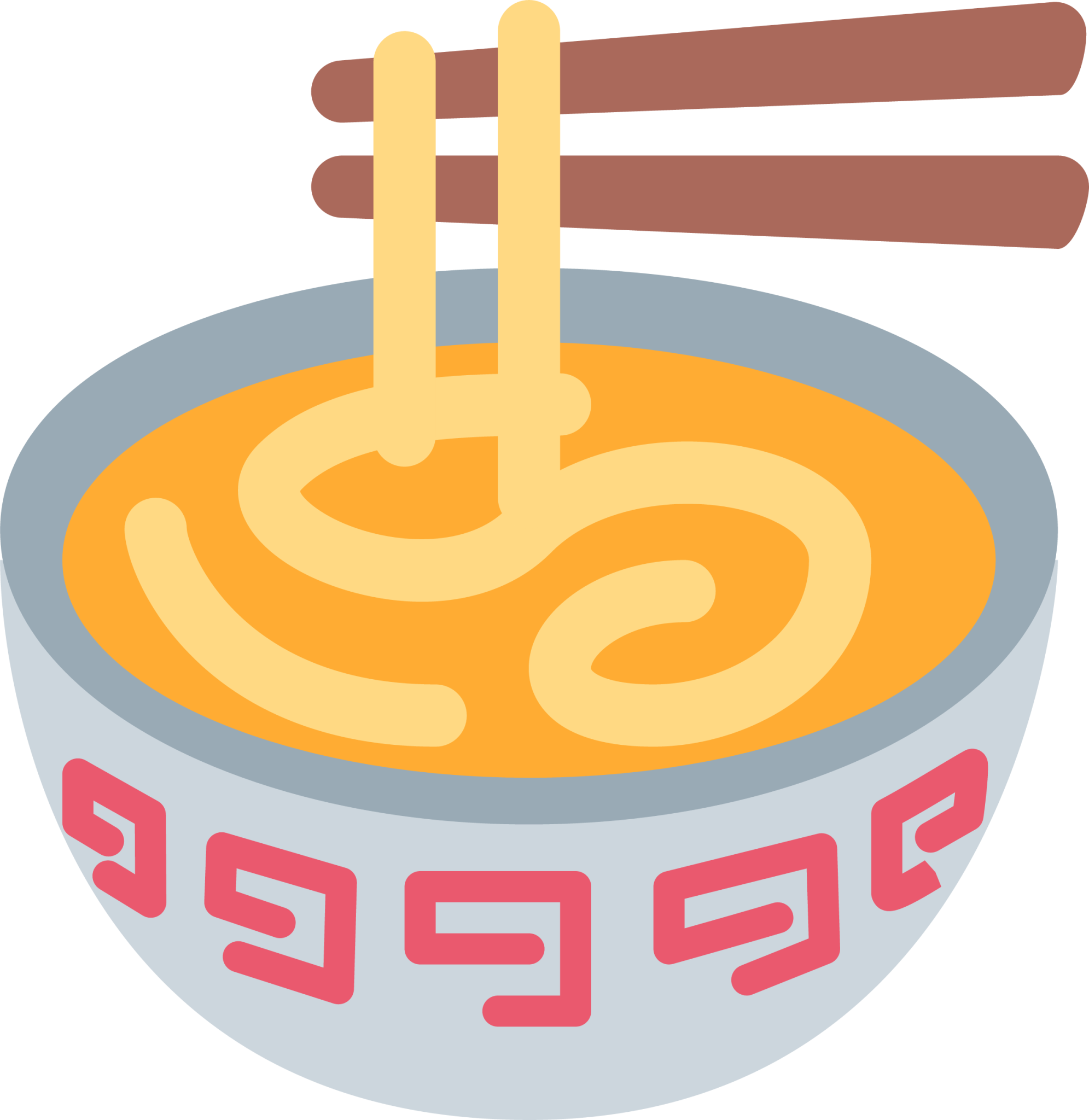}}{X}}
\usepackage{booktabs}
\usepackage{tablefootnote}
\usepackage{floatrow}
\usepackage{graphicx}
\usepackage{caption}
\usepackage{booktabs}
\usepackage[dvipsnames]{xcolor}
\usepackage{colortbl}
\usepackage{multirow}
\usepackage{subcaption}
\usepackage{hyperref}
\usepackage{url}

\usepackage[frozencache,cachedir=.]{minted} 

\usepackage{algorithm} 
\def\D{\mathcal{D}}
\def\EE{\mathbb{E}}
\def\T{\mathcal{T}}
\def\S{\mathcal{S}}
\def\W{\mathcal{W}}

\usepackage[title]{appendix}
\usepackage{amsmath}

\DeclareMathOperator*{\argmin}{arg\,min}

\def\paperID{10105} 
\def\confName{CVPR}
\def\confYear{2024}

\title{Descriptor and Word Soups \emojione: Overcoming the Parameter Efficiency \\ Accuracy Tradeoff for Out-of-Distribution Few-shot Learning }

\author{Christopher Liao\\
Boston University\\
{\tt\small cliao25@bu.edu}
\and
Theodoros Tsiligkaridis\\
MIT Lincoln Laboratory\\
{\tt\small ttsili@ll.mit.edu}
\and
Brian Kulis\\
Boston University\\
{\tt\small bkulis@bu.edu}
}

\begin{document}

\maketitle
\vspace{-3em}
\begin{abstract}
Over the past year, a large body of multimodal research has emerged around zero-shot evaluation using GPT descriptors. These studies boost the zero-shot accuracy of pretrained VL models with an ensemble of label-specific text generated by GPT. A recent study, WaffleCLIP, demonstrated that similar zero-shot accuracy can be achieved with an ensemble of random descriptors. However, both zero-shot methods are un-trainable and consequently sub-optimal when some few-shot out-of-distribution (OOD) training data is available. 
Inspired by these prior works, we present two more flexible methods called \textbf{descriptor and word soups}, which do not require an LLM at test time and can leverage training data to increase OOD target accuracy. 
Descriptor soup greedily selects a small set of textual descriptors using generic few-shot training data, then calculates robust class embeddings using the selected descriptors. Word soup greedily assembles a chain of words in a similar manner.
Compared to existing few-shot soft prompt tuning methods, word soup requires fewer parameters by construction and less GPU memory, since it does not require backpropagation. 
Both soups outperform current published few-shot methods, even when combined with SoTA zero-shot methods, on cross-dataset and domain generalization benchmarks. Compared with SoTA prompt and descriptor ensembling methods, such as ProDA and WaffleCLIP, word soup achieves higher OOD accuracy with fewer ensemble members.
Please checkout our code: \url{github.com/Chris210634/word_soups}
\end{abstract}

\vspace{-1.0em}
\section{Introduction}
\paragraph{Problem Setting} There is extensive interest from the computer vision community for training classifiers that are robust to distribution shifts. Pioneering works in this area \citep{kang2019contrastive, zhang2019bridging, saito2019semi} focused on optimizing for simple shifts in the image distribution, such as sketch-to-real adaptation. As the topic evolved, the community proposed increasingly harder adaptation problems by eliminating some restrictive assumptions. For the \emph{domain generalization (DG)} problem \citep{zhou2022domain, HPEC:CLIP:2023}, we do not assume access to unlabeled target data; for the \emph{cross-dataset generalization (XD)} problem \citep{zhou2022coop}, we allow source and target label spaces to be different; and for the \emph{parameter efficient learning (PEFT)} problem \citep{jia2022visual, wang2022learning, ICLR:MeFOMO:2024}, we impose a tight budget on the number of parameters that can be tuned. Our work lies at the confluence of these three topics. 
Similar to CoOp \citep{zhou2022coop} and MaPLe \cite{khattak2023maple}, we do assume access to labeled few-shot generic source data, such as ImageNet. 
Since we assume nothing about the relationship between source and target datasets, this setting can be more useful in practice than strict zero-shot learning. 
In this paper, we propose two parameter efficient few-shot methods, called word and descriptor soups, that finetune vision-language (VL) models to generalize to target datasets which may contain unseen labels and/or shifts in the image distribution. Our methods achieve state-of-the-art on some benchmarks without additional gradient-based tuning, but can also improve state-of-the-art gradient-based finetuning methods with an additional diversity loss.  

\begin{figure*}[t!]
    \centering
    \includegraphics[width=0.99\textwidth]{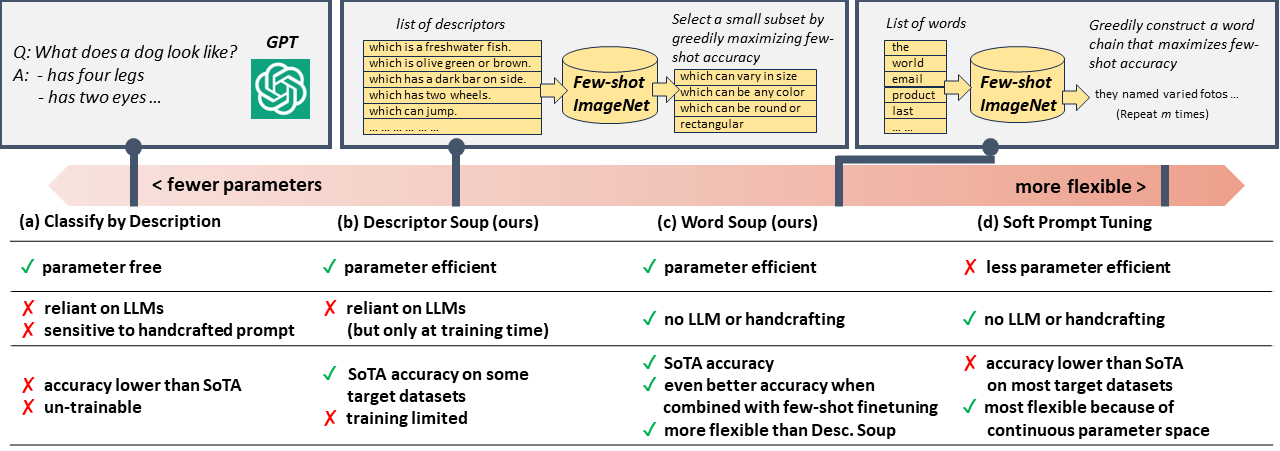}
    \caption{ Illustration of word and descriptor soups. We conceptually position our two soup methods along the tradeoff between parameter efficiency and flexibility; we then list the pros and cons of our soups compared to prior work. Firstly, word soup is more parameter efficient than soft prompt tuning, because it uses discrete tokens (see Fig. \ref{fig:parameter_efficiency}). Secondly, word soup does not require an LLM or handcrafted prompts. Lastly, word soup attains higher target accuracy than prior descriptor methods by allowing a descriptor to be any permutation of words and explicitly maximizing its accuracy on training data (see Fig. \ref{fig:descriptor_and_word_accs}). However, word soup achieves this flexibility by sacrificing the explainability of descriptors. On the other hand, descriptor soup is interpretable (see Table \ref{tab:descriptor_examples}), but less flexible than word soup, since it is limited to selecting from the pool of GPT descriptors. \vspace{-1.0em} }
    \label{fig:main}
\end{figure*}

\vspace{-1.0em}
\paragraph{Motivation} Our work is motivated by the recent success of classification by description methods \citep{menon2022visual, pratt2023does, kaul2023multi} in both zero-shot (ZS) classification and open-vocabulary object detection. These methods ask an LLM like GPT to generate a list of short descriptions for each class, then aggregate predictions from the descriptions to improve ZS accuracy, see Fig. \ref{fig:main}(a). 
It is often claimed that the impressive gain in ZS accuracy comes from additional information given by the GPT descriptions. However, a recent study called WaffleCLIP \citep{roth2023waffling} observed that random descriptors or even strings of random words can achieve similar ZS accuracy to GPT descriptors, when ensembled together (see Fig. \ref{fig:m_ablations_pretrained}). 
Therefore, gains in ZS accuracy achieved by descriptor methods are mostly driven by ensembling rather than the content of the descriptors themselves. 
Inspired by this observation, we propose descriptor and word soups, two methods which outperform WaffleCLIP by selecting descriptors or chains of words that maximize few-shot accuracy.
Word soup has 3 advantages: (1) it outperforms existing descriptor-inspired ZS methods in the few-shot OOD setting since it directly maximizes classification accuracy (see Fig. \ref{fig:main}(c)); (2) it is more parameter efficient than existing few-shot methods since the model is frozen and only the discrete descriptor tokens need to be stored; and (3) it does not require an LLM. 
The pros and cons of both descriptor and word soups are concisely stated in Figure \ref{fig:main} and discussed more in the method section.

\vspace{-1.0em}
\paragraph{Method Overview} According to the above motivation, we design a progression of three methods: \emph{descriptor soup}, \emph{word soup}, and \emph{word soup training with diversity loss}. These methods build upon each other but can be used independently and in combination with prior methods. We opted for this style of presentation, since there are motivating empirical insights at each stage, and each method achieves state-of-the-art depending on resource constraints (such as availability of an LLM at training time or parameter storage budget). 
Descriptor soup is loosely inspired by model soups \cite{wortsman2022model}; ``soup'' refers to a set of descriptors. We calculate an aggregate prediction based on the centroid of descriptors in the soup. 
We start with the most accurate descriptor on the training data and greedily add descriptors to the soup if training accuracy increases, see Fig. \ref{fig:main}(b). Similarly, for word soups, we assemble a chain of words by greedily appending a word if it increases the training accuracy of the word chain, see Fig. \ref{fig:main}(c). Finally, we present a diversity loss that can be used to optimize the CLIP model, using the word soup as an initialization. This loss is required to maintain the initial diversity among word soup members throughout finetuning.

\vspace{-1.0em}
\paragraph{Contributions} We make the following contributions to the computer vision literature:
\begin{itemize}
    \item We present word soup, which improves SoTA on few-shot cross-dataset (XD) and domain-generalization (DG) benchmarks by 1\% and 0.8\% resp.
    \item Our word soup uses fewer parameters than SoTA parameter efficient methods while achieving higher accuracy than parameter-free ZS methods in both few-shot settings.
    \item We propose a diversity loss to train VL models initialized with word soup. This allows our method to seamlessly combine with prior few-shot finetuning methods.
    \item We present qualitative results (e.g. Tab. \ref{tab:descriptor_examples}) to understand what is means for a descriptor to be ``good'', and analyze the generalizability of these descriptors (Fig. \ref{fig:descriptor_and_word_accs}). These results extend the current understanding of how descriptor and prompting methods work.
\end{itemize}

\begin{figure*}
\floatbox[{\capbeside\thisfloatsetup{capbesideposition={right,top},capbesidewidth=0.25\textwidth}}]{figure}[\FBwidth]
{\caption{Comparison with PEFT and ZS methods. 
We vary $m$ for word soup as in Fig. \ref{fig:m_ablations_pretrained}. We vary the number of prompt tokens for CoOp, VPT and MaPLe, the number of prompts for ProDA, the rank for LoRA and adapters, and the number of layers tuned for SSF and bitfit.
CoOp stores 512 parameters per soft token, while word soup stores 1 parameter per discrete token. Average of 3 runs. 
Word soup achieves the maximal CoOp accuracy with only 1/25 of the parameters on the XD benchmark and 1/70 of the parameters on the DG benchmark. Detailed results see Tab. \ref{tab:parameter_efficiency_detailed} in the Appendix. }
\label{fig:parameter_efficiency}}
{\includegraphics[width=0.7\textwidth]{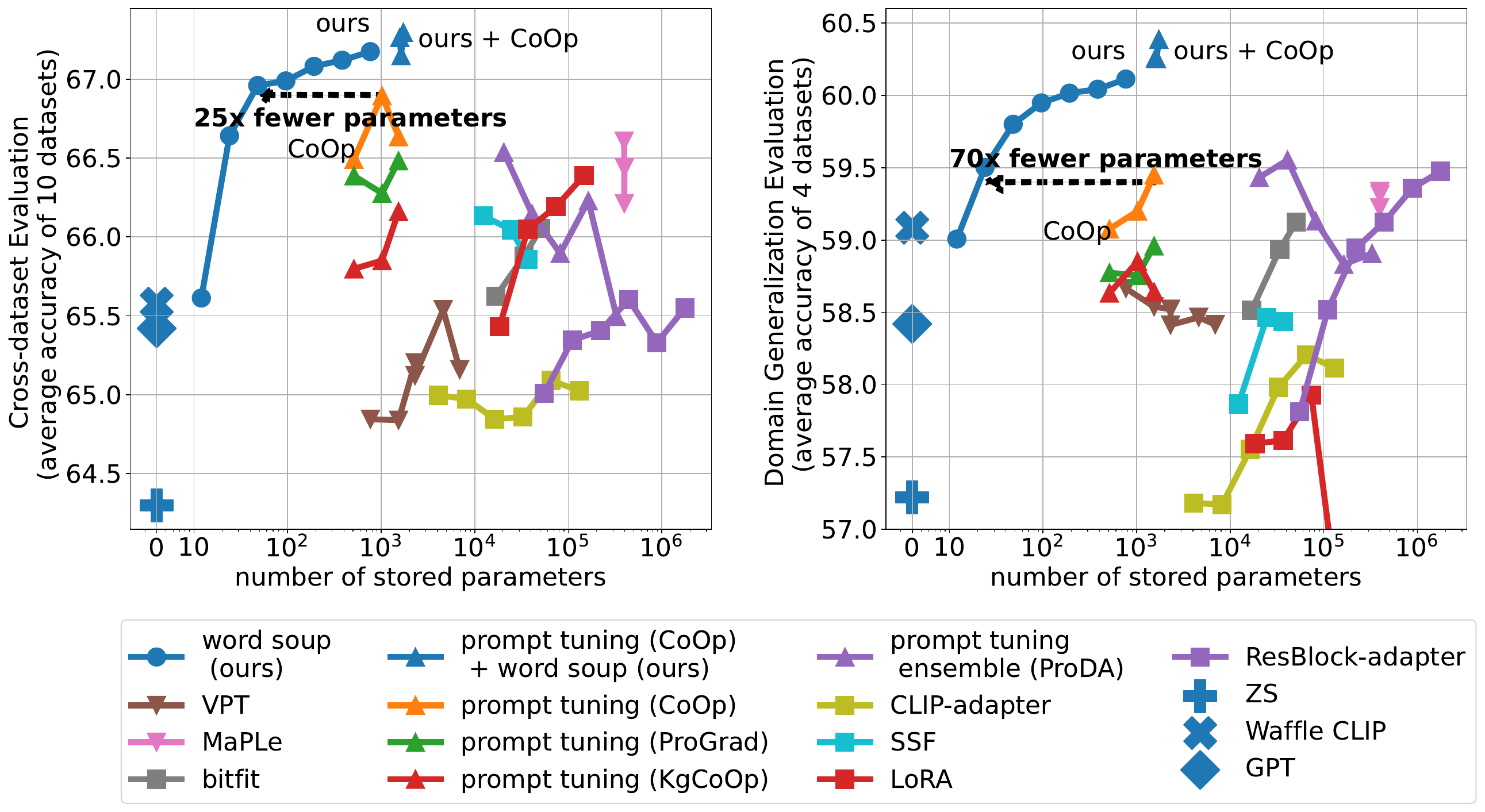}}
\vspace{-1.0em}
\end{figure*}

\vspace{-1em}
\section{Related Work}
\paragraph{Few-shot CLIP finetuning} We follow the problem settings of CoOp \citep{zhou2022coop}, CoCoOp \citep{zhou2022conditional}, MaPLe \citep{khattak2023maple}, and Clipood \citep{shu2023clipood}, which finetune a CLIP-like model \citep{radford2021learning} on few-shot ImageNet in a manner that generalizes to OOD target datasets. Many prompt tuning methods build on top of CoOp by using different loss functions \cite{chen2022prompt, yao2023visual, bulat2023lasp, derakhshani2023bayesian, peng2023sgva}, using clever optimization techniques \cite{zhu2023prompt}, ensembling multiple prompts \cite{lu2022prompt, cho2023promptstyler}, leveraging different sources of information \cite{shi2023logoprompt, he2022cpl, kan2023knowledge}, leveraging synergy between modalities \cite{zang2022unified, lin2023multimodality, khattak2023self}, or using different network architectures \cite{xing2022class, das2023learning}. We take a fundamentally different approach from these prior methods, drawing inspiration from classification by description \citep{menon2022visual}. Specifically, prior methods tune a soft prompt while our method tunes a sequence of discrete tokens.

\vspace{-1.0em}
\paragraph{Zero-shot CLIP}  Many recent papers use LLM descriptors to aid ZS or open-vocabulary visual tasks, including classification \citep{menon2022visual, pratt2023does} and detection \citep{kaul2023multi}. WaffleCLIP \citep{roth2023waffling} observed that the impressive gains in accuracy reported by these works are mostly driven by ensembling and dataset-level concepts. WaffleCLIP ensembles random descriptors and uses an LLM to discover dataset-level concepts, while we design an optimization procedure to learn good descriptors from data. Our algorithm is loosely related to model averaging methods \citep{wortsman2022model, wortsman2022robust}. However, unlike model soups \cite{wortsman2022model}, we do not generate multiple training trajectories, since all descriptors share the same model weights. 
ZS accuracy can also be improved with hierarchical label sets \cite{novack2023chils} or handcrafted prompts \cite{allingham2023simple}.
Test-time prompt tuning methods \cite{shu2022test, feng2023diverse, ma2023swapprompt, samadh2023align} train a sample-specific prompt that maximizes agreement between predictions based on a set of image augmentations. These methods suffer from long inference times due to test-time optimization.

\vspace{-1.0em}
\paragraph{Parameter efficient finetuning (PEFT)} Our word soup can be considered a PEFT \citep{hu2021lora, jia2022visual} method, but specialised to finetuning VL models in the OOD setting. Prior PEFT methods include shallow text prompt tuning \cite{zhou2022coop,zhu2023prompt,kan2023knowledge,lu2022prompt}, visual prompt tuning \cite{jia2022visual}, bias tuning \cite{zaken2021bitfit}, adapters \cite{houlsby2019parameter, gao2023clip, zhang2021tip, sung2022vl, pantazis2022svl}, LoRA \cite{hu2021lora}, SSF \cite{lian2022scaling}, side-tuning \cite{sung2022lst}, and others \cite{jie2022convolutional, yu2023task, luo2023towards, hu2023vl}. Unlike the above works, our word soup tunes fewer parameters by leveraging discrete text tokens. Similar to LST \cite{sung2022lst}, we use minimal GPU memory, since no backpropagation is required. We empirically compare with a representative subset of PEFT methods in the OOD settings in Fig. \ref{fig:parameter_efficiency}. Clearly, our word soup establishes a better tradeoff between parameter efficiency and OOD accuracy, compared to prior work.

\vspace{-0.3em}
\section {Method}
\vspace{-0.5em}
This section is organized into 4 parts. Section 3.1 reviews the classification by description \citep{menon2022visual} and WaffleCLIP \citep{roth2023waffling} methods, which motivate our soup methods. Section 3.2 presents descriptor soup, a novel intermediary method which still uses GPT descriptors at training time but not at test time. Section 3.3 presents word soup, which is similarly motivated but only requires a list of English words at training time. 
Section 3.4 describes the diversity loss used to finetune the CLIP model using word soup as the initialization. Please use Fig. \ref{fig:main} as a reference. We organize the methods in this section in order of increasing flexibility, since it is more natural to motivate word soups this way. 
However, word soups can also be motivated in the opposite direction by shortcomings of soft prompt tuning, as noted in Fig. \ref{fig:main}; this motivation is included in Appendix B. We also propose a token offset trick in Appendix C to augment descriptor soups.

\setlength\tabcolsep{4 pt}
\begin{table}[htp]
\centering
\scriptsize
\setlength\tabcolsep{6 pt}
\begin{tabular}{l l l }
\multicolumn{3}{l}{Color-coded by \textbf{source}: \colorbox{cyan}{ImageNet}, \colorbox{pink}{Pets}, \colorbox{yellow}{DTD}, \colorbox{lightgray}{Random} } \\
\toprule
 \textbf{Target}: ImageNet & Alignment & Accuracy \\
\midrule
 no descriptor                     & 0.301 & 67.1 \\
 \rowcolor{cyan}
 {which typically brightly colored.} & 0.305 (\textcolor{blue}{+0.004}) & 68.2 (\textcolor{blue}{+1.1})\\
 \rowcolor{cyan}
 which has usually white or off-white. & 0.310 (\textcolor{blue}{+0.009})  & 68.4 (\textcolor{blue}{+1.3})\\
 \rowcolor{cyan}
 which is a long, low-slung body. & 0.312 (\textcolor{blue}{+0.011})  & 68.3 (\textcolor{blue}{+1.2})\\
 \rowcolor{cyan}
 which is a curved or rectangular shape. & 0.309 (\textcolor{blue}{+0.008}) & 68.6 (\textcolor{blue}{+1.5})\\
 \rowcolor{cyan}
 which can vary in size from small to large. & 0.315 (\textcolor{blue}{+0.014}) & 68.5 (\textcolor{blue}{+1.4})\\
 \rowcolor{lightgray}
 which has reddish brown fur. & 0.300 (\textcolor{red}{-0.001}) & 66.2 (\textcolor{red}{-0.9})\\
  \rowcolor{lightgray}
 which is a hard skeleton. & 0.295 (\textcolor{red}{-0.006}) & 66.6 (\textcolor{red}{-0.5})\\
 \rowcolor{pink}
which is a medium-sized, short-haired cat. & 0.291 (\textcolor{red}{-0.010}) & 66.0 (\textcolor{red}{-1.1})\\
\rowcolor{pink}
which has sharp claws. & 0.299 (\textcolor{red}{-0.002}) & 66.6 (\textcolor{red}{-0.5})\\
\rowcolor{yellow}
which is a repeating pattern. & 0.295 (\textcolor{red}{-0.006}) & 66.1 (\textcolor{red}{-1.0})\\
\rowcolor{yellow}
which is a sign with the shop’s name. & 0.295 (\textcolor{red}{-0.006}) & 66.7 (\textcolor{red}{-0.4})\\
\bottomrule
\end{tabular}
\setlength\tabcolsep{4 pt}
\begin{tabular}{lll}
\toprule
 \textbf{Target}: Pets  & Alignment & Accuracy \\
\midrule
no descriptor & 0.322 & 88.4 \\
a type of pet. (handcrafted; for reference) & 0.331 (\textcolor{blue}{+0.009}) & 89.0 (\textcolor{blue}{+0.6})\\
\rowcolor{pink}
which is a large, powerful cat. & 0.321 (\textcolor{red}{-0.001}) & 89.8 (\textcolor{blue}{+1.4})\\
\rowcolor{pink}
which has sharp claws. & 0.324 (\textcolor{blue}{+0.002}) & 89.9 (\textcolor{blue}{+1.5})\\
\rowcolor{pink}
which has soulful eyes. & 0.317 (\textcolor{red}{-0.005}) & 89.9 (\textcolor{blue}{+1.5})\\
\rowcolor{pink}
which is a long arm with a claw ... & 0.324 (\textcolor{blue}{+0.002}) & 87.8 (\textcolor{red}{-0.6})\\
\rowcolor{pink}
which is a medium-sized, short-haired cat. & 0.327 (\textcolor{blue}{+0.005}) & 91.4 (\textcolor{blue}{+3.0})\\
\rowcolor{lightgray}
which is a boat with sails. & 0.293 (\textcolor{red}{-0.029}) & 81.5 (\textcolor{red}{-6.9})\\
\rowcolor{lightgray}
which often used by knights and soldiers. & 0.315 (\textcolor{red}{-0.007}) & 80.8 (\textcolor{red}{-7.6})\\
\rowcolor{cyan}
which can vary in size from small to large. & 0.333 (\textcolor{blue}{+0.011}) & 88.6 (\textcolor{blue}{+0.2})\\
\rowcolor{cyan}
which typically has a yellow or brownish color. & 0.335 (\textcolor{blue}{+0.013}) & 89.3 (\textcolor{blue}{+0.9})\\
\bottomrule
\end{tabular}
\begin{tabular}{lll}
\toprule
 \textbf{Target}: Textures (DTD)  & Alignment & Accuracy \\
\midrule
no descriptor & 0.273 & 44.3 \\
a type of texture. (handcrafted; for reference) & 0.287 (\textcolor{blue}{+0.014}) & 44.1 (\textcolor{red}{-0.2})\\
\rowcolor{yellow}
which may be decorated with a pattern or logo. & 0.286 (\textcolor{blue}{+0.013}) & 47.2 (\textcolor{blue}{+2.9})\\
\rowcolor{yellow}
which is a sign with the shop's name. & 0.261 (\textcolor{red}{-0.012}) & 45.3 (\textcolor{blue}{+1.0})\\
\rowcolor{yellow}
which is a backdrop. & 0.280 (\textcolor{blue}{+0.007}) & 46.6 (\textcolor{blue}{+2.3})\\
\rowcolor{yellow}
which is a repeating pattern. & 0.283 (\textcolor{blue}{+0.010}) & 46.3 (\textcolor{blue}{+2.0})\\
\rowcolor{yellow}
which typically has a pattern or design. & 0.295 (\textcolor{blue}{+0.022}) & 45.5 (\textcolor{blue}{+1.2})\\
\rowcolor{lightgray}
which is a guard tower. & 0.243 (\textcolor{red}{-0.030}) & 43.4 (\textcolor{red}{-0.9})\\
\rowcolor{lightgray}
which has loud crow. & 0.253 (\textcolor{red}{-0.020}) & 42.4 (\textcolor{red}{-1.9})\\
\rowcolor{cyan}
which can be brightly colored or patterned. & 0.283 (\textcolor{blue}{+0.010}) & 44.5 (\textcolor{blue}{+0.2})\\
\rowcolor{cyan}
which is a curved or rectangular shape. & 0.281 (\textcolor{blue}{+0.008}) & 44.4 (\textcolor{blue}{+0.1})\\
\bottomrule
\end{tabular}
\caption{Qualitative comparison of descriptors. We select descriptors based on a source dataset using Alg. 1 and test on a target dataset. The tables are organized by the target dataset; the color of the highlight indicates the source dataset. We include randomly selected descriptors in gray for comparison. 
Alignment refers to the average cosine similarity between image embeddings and the corresponding text embeddings. 
Observe that selected descriptors tend to describe the source dataset as a whole and improve both accuracy and alignment.
Also observe that a descriptor soup trained on ImageNet (blue) generalizes to other datasets, but not vice versa.
\vspace{-1.0em}}
\label{tab:descriptor_examples}
\end{table}
\setlength\tabcolsep{6 pt}

\setlength\tabcolsep{4 pt}
\begin{table}[htp]
\centering
\scriptsize
\begin{tabular}{llll}
\toprule
 \textbf{Target}: ImageNet  & Alignment & Uniformity & Accuracy \\
\midrule
 no descriptor                     & 0.301 & 0.173 & 67.1 \\
 \rowcolor{cyan}
dat they ... difficulties. & 0.306 (\textcolor{blue}{+0.005}) & 0.174 ({+0.001}) & 68.9 (\textcolor{blue}{+1.8})\\
\rowcolor{cyan}
 similar vary ... mention etc. & 0.314 (\textcolor{blue}{+0.013}) & 0.183 ({+0.010}) & 69.1 (\textcolor{blue}{+2.0})\\
 \rowcolor{cyan}
 separately aspects ... adopted. & 0.315 (\textcolor{blue}{+0.014}) & 0.181 ({+0.008}) & 69.2 (\textcolor{blue}{+2.1})\\
 \rowcolor{cyan}
 tue alot ... itself. & 0.303 (\textcolor{blue}{+0.002}) & 0.178 ({+0.005}) & 69.0 (\textcolor{blue}{+1.9})\\
 \rowcolor{cyan}
 bufing beginner ... status. & 0.311 (\textcolor{blue}{+0.010}) & 0.181 ({+0.008}) & 68.8 (\textcolor{blue}{+1.7})\\
 \rowcolor{lightgray}
soviet vbulletin ... inexpensive. & 0.320 (\textcolor{blue}{+0.019}) & 0.195 (\textcolor{red}{+0.022}) & 62.0 (\textcolor{red}{-5.1})\\
\rowcolor{lightgray}
 ideal ips ... filename. & 0.314 (\textcolor{blue}{+0.013}) & 0.196 (\textcolor{red}{+0.023}) & 59.7 (\textcolor{red}{-7.4})\\
\bottomrule
\end{tabular}
\caption{Example of a 5 member word soup trained on ImageNet (in blue) along with random chains of words (in gray) for comparison. Comparing with Tab. \ref{tab:descriptor_examples}, we observe that the word soup descriptors achieve higher accuracy than descriptor soups, since word soup is more flexible from an optimization perspective. Here, we include uniformity scores, since chains of random words improve alignment at the expense of increasing uniformity.
Uniformity is the average cosine similarity between image and text embeddings with different labels. \vspace{-1.0em}}
\label{tab:wordsoup_examples}
\end{table}
\setlength\tabcolsep{6 pt}


\subsection{LLM Descriptors and WaffleCLIP}
Several works use LLM descriptors to supplement class names in VL models \citep{menon2022visual, pratt2023does, kaul2023multi}. These methods ask an LLM to describe the object being classified and incorporate this information into the textual input by forming sentences such as ``a photo of a tench, which is a freshwater fish'' or ``a photo of a goldfish, which has small black eyes''. The LLM generates on average 5.8 such descriptors per label, and the centroids of the resulting text embeddings are used for zero-shot classification of images. The improvement in zero-shot accuracy can be attributed to (1) additional information coming from the LLM and (2) ensembling. In WaffleCLIP, \citet{roth2023waffling} claim that most of the gain in accuracy reported by \citet{menon2022visual} can be attributed to ensembling. They showed that appending a similar number of \emph{randomly selected} descriptors to the class names can  achieve similar zero-shot accuracies as the GPT descriptors. We confirm this result in Fig. \ref{fig:m_ablations_pretrained}. Observe in this figure that both random descriptors (labeled as ``random soup'') and chains of random nonsensical words (labeled as ``waffle CLIP'') perform better than classification by description (``GPT centroids'') for the same number of descriptors per label ($m$). This is a surprising result. We reason that selecting descriptors which maximize few-shot training accuracy would achieve higher accuracy than random descriptors; this motivates descriptor soup.

\setlength\tabcolsep{4.5 pt}
\begin{table*}[t!]
\scriptsize
\centering
\begin{tabular}{l c ccc ccc ccccc >{\columncolor[gray]{0.9}}ccccc >{\columncolor[gray]{0.9}}c}
\toprule
&& \textbf{Source} & \multicolumn{11}{c}{\textbf{Cross-dataset (XD) Evaluation Targets}} & \multicolumn{5}{c}{\textbf{Domain Generalization Targets}} \\

\cmidrule(lr){3-3} \cmidrule(lr){4-14} \cmidrule(lr){15-19}

     & $m$ & \rotatebox{90}{ INet } & \rotatebox{90}{ Caltech } & \rotatebox{90}{ Pets } & \rotatebox{90}{ Cars } & \rotatebox{90}{ Flowers } & \rotatebox{90}{ Food } & \rotatebox{90}{ Aircraft } & \rotatebox{90}{ SUN } & \rotatebox{90}{ DTD } & \rotatebox{90}{ EuroSAT } & \rotatebox{90}{ UCF } & \rotatebox{90}{ \textbf{Mean} } & \rotatebox{90}{ INet-V2 } & \rotatebox{90}{ Sketch } & \rotatebox{90}{ INet-A } & \rotatebox{90}{ INet-R } & \rotatebox{90}{ \textbf{Mean} }\\
     
    \midrule
    
     CLIP ZS \citep{zhou2022coop} &      1 &  67.1 & 93.3 & 89.0 & 65.4 & 71.0 & 85.7 & 25.0 & 63.2 & 43.6 & 46.7 & 67.4 & 65.02 & 61.0 & 46.6 & 47.2 & 74.1 & 57.22 \\
     Ensemble \citep{radford2021learning} &      80 &  68.4 & 93.5 & 88.8 & 66.0 & 71.1 & 86.0 & 24.8 & 66.0 & 43.9 & 45.0 & 68.0 & 65.31 & 61.9 & 48.5 & 49.2 & \textbf{77.9} & 59.36 \\
     GPT centroids \citep{menon2022visual} &      5.8 &  68.2 & 94.1 & 88.4 & 65.8 & 71.5 & 85.7 & 24.7 & \textbf{67.5} & 44.7 & 46.6 & 67.4 & 65.63 & 61.5 & 48.2 & 48.9 & 75.1 & 58.40 \\
     GPT score mean \citep{menon2022visual} &      5.8 &  68.6 & 93.7 & 89.0 & 65.1 & 72.1 & 85.7 & 23.9 & 67.4 & 44.0 & 46.4 & 66.8 & 65.42 & 61.8 & 48.1 & 48.6 & 75.2 & 58.42 \\
     \midrule
     Random descriptors &      16 &  67.9 & 94.1 & 87.6 & 65.6 & 71.5 & 85.6 & 24.9 & 66.1 & 44.7 & 49.1 & 67.2 & 65.65 & 61.6 & 48.7 & 50.0 & 76.7 & 59.22 \\
     {  $\quad$ + offset trick (ours)} &      96 &  68.5 & 93.5 & 89.2 & 65.8 & 72.0 & 85.7 & 25.2 & 66.1 & 44.4 & 53.0 & 68.2 & 66.29 & 61.9 & 48.9 & \textbf{50.6} & 77.5 & 59.76 \\
     Waffle CLIP \citep{roth2023waffling} &      16 &  68.1 & 93.5 & 88.4 & 65.4 & 72.0 & 85.9 & 25.9 & 66.2 & 44.1 & 46.3 & 68.0 & 65.58 & 61.8 & 48.6 & 49.8 & 76.2 & 59.08 \\
     { $\quad$  + offset trick (ours)} &      96 &  68.6 & 93.1 & 89.5 & 65.9 & 72.1 & 86.1 & \textbf{26.3} & 66.2 & 44.2 & 52.5 & 68.8 & 66.49 & 62.1 & 48.9 & 50.2 & 77.1 & 59.59 \\
     Descriptor soup (ours) &      16.7 &  68.9 & \textbf{94.7} & 89.4 & \textbf{66.2} & 72.2 & 86.2 & 25.5 & 67.3 & 45.1 & 46.6 & 68.7 & 66.18 & 62.1 & 48.7 & 49.7 & 76.4 & 59.25 \\
     { $\quad$  + offset trick (ours) } &      100 &  69.1 & 93.8 & \textbf{89.8} & 66.0 & \textbf{72.9} & \textbf{86.2} & 25.4 & 66.8 & 45.0 & 51.6 & \textbf{69.1} & 66.67 & 62.6 & \textbf{49.0} & 50.5 & 77.2 & 59.82 \\
     \midrule
     Word soup (ours) &      8 &  69.2 & 94.4 & 89.5 & 65.4 & 72.3 & 85.8 & 25.8 & 67.4 & 44.7 & 53.5 & 68.4 & 66.72 & 62.9 & 48.7 & 50.2 & 77.0 & 59.69 \\
     Word soup score mean (ours) &      8 &  \textbf{69.4} & 94.3 & 89.6 & 65.4 & 72.4 & 85.9 & 25.9 & 67.3 & \textbf{45.2} & \textbf{55.8} & 68.5 & \textbf{67.03} & \textbf{63.0} & 49.0 & 50.4 & 77.2 & \textbf{59.90} \\
     {  $\quad$ gain over GPT} &       &  \textcolor{blue}{+0.8} & \textcolor{blue}{+0.6} & \textcolor{blue}{+0.6} & \textcolor{blue}{+0.3} & \textcolor{blue}{+0.3} & \textcolor{blue}{+0.2} & \textcolor{blue}{+2.0} & \textcolor{red}{-0.1} & \textcolor{blue}{+1.2} & \textcolor{blue}{+9.4} & \textcolor{blue}{+1.7} & \textcolor{blue}{+1.6} & \textcolor{blue}{+1.2} & \textcolor{blue}{+0.9} & \textcolor{blue}{+1.8} & \textcolor{blue}{+2.0} & \textcolor{blue}{+1.5} \\
     { $\quad$ gain over Waffle} &       &  \textcolor{blue}{+1.3} & \textcolor{blue}{+0.8} & \textcolor{blue}{+1.2} & \textcolor{blue}{+0.0} & \textcolor{blue}{+0.4} & \textcolor{blue}{+0.0} & \textcolor{red}{-0.0} & \textcolor{blue}{+1.1} & \textcolor{blue}{+1.1} & \textcolor{blue}{+9.5} & \textcolor{blue}{+0.5} & \textcolor{blue}{+1.5} & \textcolor{blue}{+1.2} & \textcolor{blue}{+0.4} & \textcolor{blue}{+0.6} & \textcolor{blue}{+1.0} & \textcolor{blue}{+0.8} \\
\bottomrule
  \end{tabular}
  \caption{ Comparison with ZS methods. All baseline methods in this table use prompts/descriptors on top of the pretrained model in a ZS manner. Note that the soup methods are not truly zero-shot because they require some training data. However, we do compare against all baselines in the few-shot setting in Table \ref{tab:ft}. We use the ViT/B-16 CLIP model trained by Open-AI. All non-deterministic numbers are an average of 3 random seeds. $m$ indicates the number of descriptors used. ``Ensemble'' refers to the set of 80 handcrafted prompts created by Open-AI; GPT score mean corresponds to the classification by description method. We use centroid evaluation unless ``score mean'' is explicitly stated. We achieve substantial gains over GPT descriptors and waffle CLIP as indicated in the bottom two rows. \vspace{-1.0em}}
  \label{tab:pretrained}
\end{table*}
\setlength\tabcolsep{6 pt}

\vspace{-0.5em}
\subsection{Descriptor Soup}
\vspace{-0.5em}
We reference Alg. 1 in the Appendix throughout this section. Let $\D = \{d_1, ..., d_n\}$ denote a set of $n$ descriptors such as ``which is a freshwater fish''. These descriptors are obtained by combining all descriptors generated by GPT for 1,000 ImageNet classes \cite{menon2022visual}, and keeping only unique entries. Descriptors are no longer connected to their original classes. We wish to select a set of $m$ descriptors that maximizes accuracy on few-shot training data. Let's define the loss function $\ell(\S_{\text{train}}, \T_{\text{train}}(d))$ to be the 0-1 loss of the model using descriptor $d$ over the entire training dataset $\S_{\text{train}}$. $\T_{\text{train}}(d)$ denotes the label text embeddings calculated by the text encoder by appending descriptor $d$ to all class names. Since all parameters of the vision model remain constant, we ignore vision model parameters in the notation. We aim to find a set of $m$ descriptors whose centroids in the text embedding space minimize the 0-1 loss:
\vspace{-0.5em}
\begin{equation}
    \D^*_m = \{d_1^*, ..., d_m^*\} = \argmin_{d_{1:m} \in \D} \ell\left(\S_{\text{train}}, \frac{1}{m} \sum_{i=1}^m \T_{\text{train}}(d_i) \right)
    \label{eq:descriptor_soup}
\end{equation}
Note that $\frac{1}{m} \sum_{i=1}^m \T_{\text{train}}(d_i)$ denotes the L2-normalized centroid of text embeddings for each class. We always normalize the centroid so it can be used to calculate the cosine similarity with image embeddings; this is omitted from the math to avoid clutter.

Eq. \ref{eq:descriptor_soup} is an intractable combinatorial problem, but we can approximately solve it via a greedy approach or by solving the continuous version of the problem using gradient descent. We use a greedy approach, inspired by \citet{wortsman2022model}. The algorithm can be summarized as (reference Alg. 1):
\begin{enumerate}
    \item Calculate $\ell(\S_{\text{train}}, \T_{\text{train}}(d))$ for all $d \in \D$. Sort the descriptors by increasing loss / decreasing accuracy. With slight abuse of notation, denote the sorted list as $\D = [d_0, ..., d_n]$.
    \item Initialize the ``descriptor soup'' $\D^* = \{d_0\}$ with the best descriptor.
    \item For $i$ in $1:n$: Add $d_i$ to $\D^*$ if it decreases the loss of $\D^*$.
    \item Return the first $m$ descriptors in $\D^*$.
\end{enumerate}
Please find ZS results for descriptor soup in Tab. \ref{tab:pretrained}.

\vspace{-0.5em}
\paragraph{Building Intuition} A natural question to ask is: descriptor soup members no longer describe individual classes, so why does Alg. 1 work? The answer has two parts (1) Alg. 1 finds \emph{descriptors which describe the dataset as a whole, rather than individual labels}; these descriptors are orthogonal to the classification problem and \emph{increase classification accuracy by increasing alignment between corresponding image and text embeddings.} (2) Descriptor soups generalize when the target classification problem has a narrower scope than the source classification problem.
Prior work (e.g. \citep{zhou2022coop, khattak2023maple, roth2023waffling}) suggests that handcrafted dataset-specific descriptors such as ``a type of aircraft'' or ``a type of pet'' improve ZS accuracy. Dataset-level descriptors like these are easier to design than label-level descriptors, so using dataset-level descriptors is currently standard practice. We hypothesize that these descriptors improve accuracy by increasing alignment between corresponding image and text embeddings; we demonstrate this in Tab. \ref{tab:descriptor_examples}. e.g. ``a type of pet'' improves pet classification accuracy by 0.6\% and alignment by 0.01. 

We further hypothesize that descriptor soup members learn to mimic the behavior of handcrafted dataset-level descriptors. We display examples of descriptor soups trained on three different datasets in Table \ref{tab:descriptor_examples} in support of this intuition. 
Descriptors trained on pets (in pink) mention ``claws'', ``eyes'', and ``hair'', which are concepts common to most pets. In a similar vein, descriptors trained on textures/DTD (in yellow) mention ``pattern'', ``logo'', and ``design''. 
Meanwhile, ImageNet is a broader dataset, so descriptors trained on ImageNet (in blue) are generally non-specific (e.g. ``which could be brown or grey''). This is intuitive, since ImageNet is a dataset with diverse classes. A descriptor such as ``which is a type of dog'' would be detrimental to the zero-shot accuracy, since it would bias the classifier toward labels that are types of dogs. 
Table \ref{tab:descriptor_examples} shows that individual descriptor soup members increase both the alignment and classification accuracy, when the source and target datasets are the same. The next paragraph addresses the issue of generalizability when source and target datasets are different. 

\begin{figure}[t!]
    \centering
    \includegraphics[width=1.\linewidth]{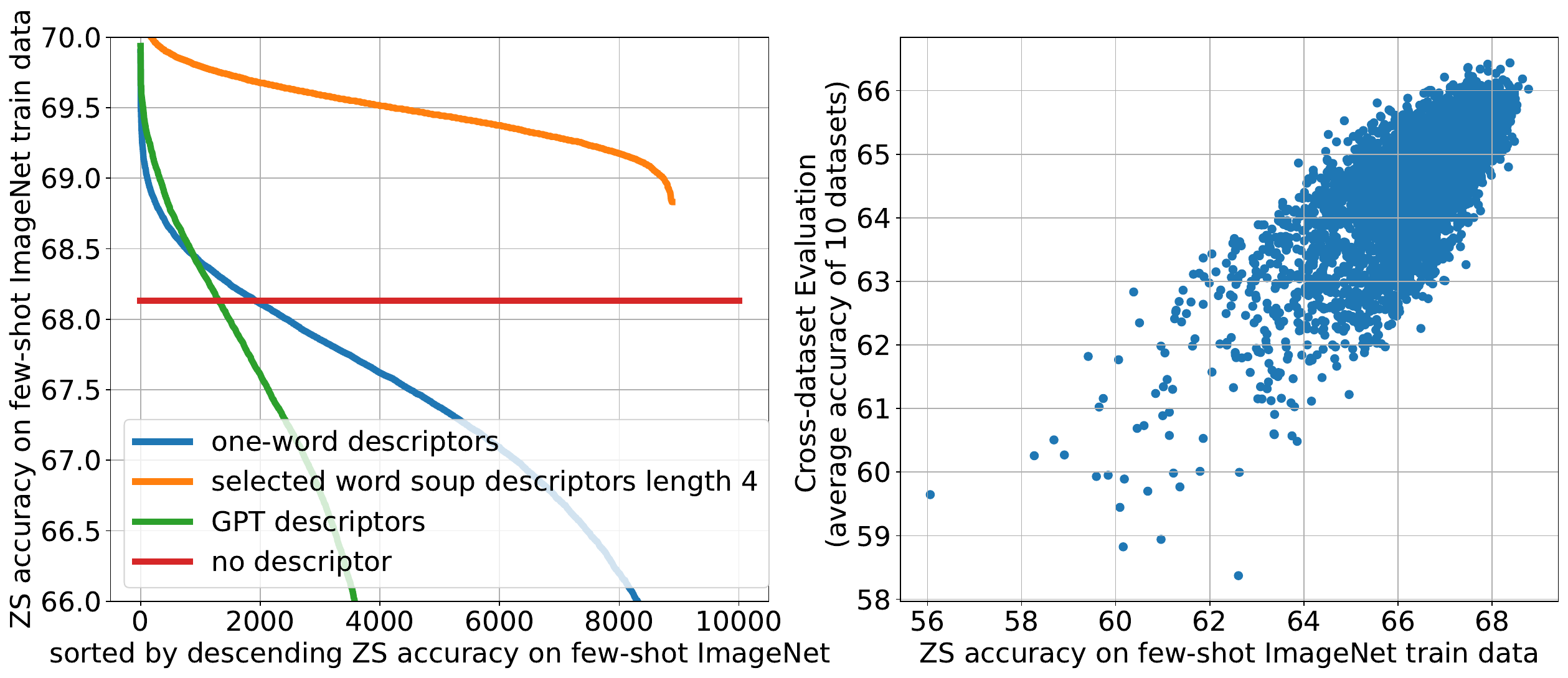}
    \caption{(Left) Plot of ImageNet accuracy when the same descriptor is appended to every class label. 
    Observe that there are more than 1,000 GPT descriptors and single-word descriptors that are better than standard ZS (in red). 
    When we further consider word chains of length 4, the number of accurate descriptors increases dramatically (orange). 
    (Right) Scatter plot of average target accuracy vs. ImageNet accuracy of GPT descriptors. We observe a positive correlation, so descriptors trained on ImageNet are likely to generalize to other datasets. 
    \vspace{-1.0em} }
    \label{fig:descriptor_and_word_accs}
\end{figure}

\vspace{-0.5em}
\paragraph{Generalizability} 
Descriptor soups trained on ImageNet generalize to target datasets with narrower scopes, but not vice versa. This is because ImageNet concepts are a superset of narrower target datasets; e.g. ImageNet classes contain types of cars and pets. Table \ref{tab:descriptor_examples} shows that descriptors trained on ImageNet (blue) improve both the alignment and accuracy on Pets and Textures; but descriptors trained on the latter two datasets (pink and yellow) decrease the same metrics on ImageNet. To further support the generalizability of descriptor soups, we show a positive correlation between ImageNet accuracy and \emph{average} target dataset accuracy in Fig. \ref{fig:descriptor_and_word_accs} (right). 
Finally, we train a descriptor soup on \emph{test data} to maximize average accuracy of 10 datasets; we call this the ``descriptor soup upper bound'' in the middle of Tab. \ref{tab:ft}. The upper bound only achieves marginal improvement over the descriptor soup trained on ImageNet (three rows above the upper bound in Tab. \ref{tab:ft}). This suggests that greedily maximizing the descriptor soup accuracy on ImageNet training data is a good approximation of maximizing the target accuracy; i.e. the generalization gap is small.

\subsection{Word Soup}
Descriptor soup achieves impressive state-of-the-art performance, but it is still \emph{reliant on an LLM} at training time to generate a list of candidate descriptors and is \emph{limited} to this fixed descriptor list.
In order to remove the reliance on LLMs and make the optimization process more flexible, we propose to generate descriptors in a greedy fashion using individual words selected from a dictionary. 
We use the list of 10,000 most commonly-used words on the web\footnote{\url{github.com/first20hours/google-10000-english}} as the candidate pool of words. 

Given a list of $n$ words $\mathcal{W} = \{w_1, ..., w_n\}$ (we abuse some notations slightly, since the word soup is a separate method). 
Descriptors are allowed to be any sequence of words, as long as the length does not exceed $p$. Concretely,
\begin{equation}
\begin{split}
    & \D^*_m = \{d_1^*, ..., d_m^*\} = \argmin_{d_{1:m} \in D' } \ell\left(\S_{\text{train}}, \frac{1}{m} \sum_{i=1}^m
    \T_{\text{train}}(d_i) \right)
    \\
    & D' := \{\text{all $q$ permutations of $\W$, $\forall q \le p$} \}
\end{split}
\label{eq:word_soup}
\end{equation}
The word soup problem described by Eq. \ref{eq:word_soup} is again intractable, so we propose an approximate greedy solution using the following steps (see Alg. 2 in the Appendix):
\begin{enumerate}
    \item \textbf{Initialization:} Sort $\mathcal{W}$ by decreasing ZS accuracy to filter out unsuitable words (see Fig. \ref{fig:descriptor_and_word_accs} left). For this step, we only consider single word descriptors (e.g. ``a photo of a cat, the.''). 
    Select the top-$k_0$ and top-$k_1$ words, denoted as $\W_{\text{top}k_0}$ and $\W_{\text{top}k_1}$, resp. $k_0 < k_1$.
    \item Randomly select a word $w$ from $\W_{\text{top}k_0}$ and initialize the descriptor $d=w$.
    \item Shuffle $\W_{\text{top}k_1}$. Then, for $w' \in \W_{\text{top}k_1}$, 
    append $w'$ to $d$, only if it increases the accuracy of $d$.
    \item return $d$.
\end{enumerate}
We obtain a total of $m$ independent (in a loose sense) descriptors by repeating steps 2-4. In these steps, we randomly select from $\W_{\text{top}k_0}$ and shuffle $\W_{\text{top}k_1}$ to encourage diversity among the $m$ selected descriptors. Instead of truncating all descriptors to a pre-determined length $p$, we introduce a patience parameter in Alg. 2, which implicitly controls the average descriptor length. We now motivate word soup.

\paragraph{Motivation from descriptor soup} The descriptor soup method has some intuitive properties covered in the previous sub-section, but is limited by the small number of good descriptors. Fig. \ref{fig:descriptor_and_word_accs} left shows that only about 1,200 descriptors (green line) in $\D$ are better than no descriptor (vanilla ZS; red line). The descriptor soup is limited to various combinations of these 1,200 ``good'' descriptors. On the contrary, when we expand the hypothesis space to be $D'$, any permutation of a set of words, there are many more good descriptors to choose from, as indicated by the orange line in Fig. \ref{fig:descriptor_and_word_accs} left. In other words, word soup improves classification accuracy by increasing the size of the hypothesis class. Tab. \ref{tab:wordsoup_examples} supports this assertion by showing that individual word soup descriptors achieve higher accuracies on ImageNet than descriptor soup members. 

\setlength\tabcolsep{3.2 pt}
\begin{table}[t!]
\scriptsize
\centering
\begin{tabular}{l c c >{\columncolor[gray]{0.9}} c >{\columncolor[gray]{0.9}}c}
\toprule

     & $m$ & Source & \textbf{XD Mean} & \textbf{DG Mean} \\
     & & INet & (10 datasets) & (4 datasets) \\
     
    \midrule

    CLIP ZS &      1 &  67.1 & 65.02 & 57.22 \\
     Vanilla CoOp &      1 &  70.0 & 66.52 & 59.25 \\
     { $\quad$ + word soup} &      8 &  69.6 & 66.59 & 59.26 \\
     CoOp ensemble &      8 &  69.8 & 66.68 & 59.18 \\
     \midrule
     
     CoOp regularized towards initialization & 1 &  70.2 & 66.97 & 59.94 \\
     { $\quad$ + word soup} & 8 &  69.9 & 66.69 & 60.05 \\
     CoOp with label smoothing &  1 &  70.1 & 66.37 & 60.09 \\
     { $\quad$ + word soup} &      8 &  69.9 & 66.13 & 60.16 \\
     CoOp + word soup ($\lambda=0$) &      8 &  69.8 & 66.21  & 59.15 \\
     {$\quad$ \textbf{+ our diversity loss} ($\lambda=0.25$)} &      8 &  \textbf{70.2} & \textbf{67.23} & \textbf{60.20} \\
    
\bottomrule
  \end{tabular}
  \caption{Ablation results to support the diversity loss. ``Vanilla CoOp + word soup'' refers to appending the word soup descriptors directly to soft CoOp prompts. ``CoOp ensemble'' refers to ensembling $m$ randomly-initialized soft descriptors trained with CoOp. 
  Observe that the model trained with our diversity loss ($\lambda=0.25$) achieves a 1\% increase in accuracy on average. This increase in accuracy cannot be achieved with label smoothing or regularization towards the initialization as in MIRO \citep{cha2022domain} and ProGrad \citep{zhu2023prompt}. Detailed results see Tab. \ref{tab:coop-appendix} in the Appendix.
   }
  \label{tab:coop}
\end{table}
\setlength\tabcolsep{6 pt}

\setlength\tabcolsep{4.2 pt}
\begin{table}[t!]
\scriptsize
\centering
\begin{tabular}{l c ccc ccc }
\toprule

&& \multicolumn{5}{c}{\textbf{Cross-dataset Evaluation Target Mean}} \\
\cmidrule(lr){3-7}

 & $m$ & B/32$\dagger$ & B/16$\dagger$ & L/14$\ddagger$ & CoCa L/14$\ddagger$ & g/14$\ddagger$ \\
 \midrule
ZS & 1 &61.32 & 65.02 & 73.11 & 74.82 & 77.58 \\
GPT score mean & 5.8 & 61.22 & 65.42 & 73.08 & 75.48 & 77.14 \\
Waffle CLIP & 16 & 62.13 & 65.58 & 73.25 & 75.37 & 77.72 \\
Desc. soup + offsets & 100 & \textbf{62.79} & 66.67 & 73.19 & 75.95 & 78.04 \\
Word soup (ours) & 8 & 62.24 & \textbf{67.03} & \textbf{73.56} & \textbf{76.08} & \textbf{78.09} \\

\midrule

&& \multicolumn{5}{c}{\textbf{Domain Generalization Evaluation Target Mean}} \\
\cmidrule(lr){3-7}

 & $m$ & B/32$\dagger$ & B/16$\dagger$ & L/14$\ddagger$ & CoCa L/14$\ddagger$ & g/14$\ddagger$ \\
 \midrule
ZS & 1 & 47.68 & 57.22 & 64.88 & 67.94 & 71.37 \\
GPT score mean & 5.8 & 47.95 & 58.42 & 64.96 & 67.67 & 71.26 \\
Waffle CLIP & 16 & 49.07 & 59.08 & 64.47 & 67.85 & 70.99 \\
Desc. soup + offsets & 100 & \textbf{50.05} & 59.82 & \textbf{65.81} & 68.32 & \textbf{72.21} \\
Word soup (ours) & 8 & 50.00 & \textbf{59.90} & 65.73 & \textbf{68.73} & 72.05 \\
    
\bottomrule
  \end{tabular}
  \caption{ Comparison with ZS baselines at different model scales. $\dagger$ indicates a model trained by Open-AI \cite{radford2021learning}; $\ddagger$ indicates a model trained by Open-CLIP \cite{ilharco_gabriel_2021_5143773}. Detailed results see Tab. \ref{tab:model-scaling-detailed} in the Appendix. }
  \label{tab:scaling}
\end{table}
\setlength\tabcolsep{6 pt}

\begin{figure}[t!]
    \centering
    \includegraphics[width=1.\linewidth]{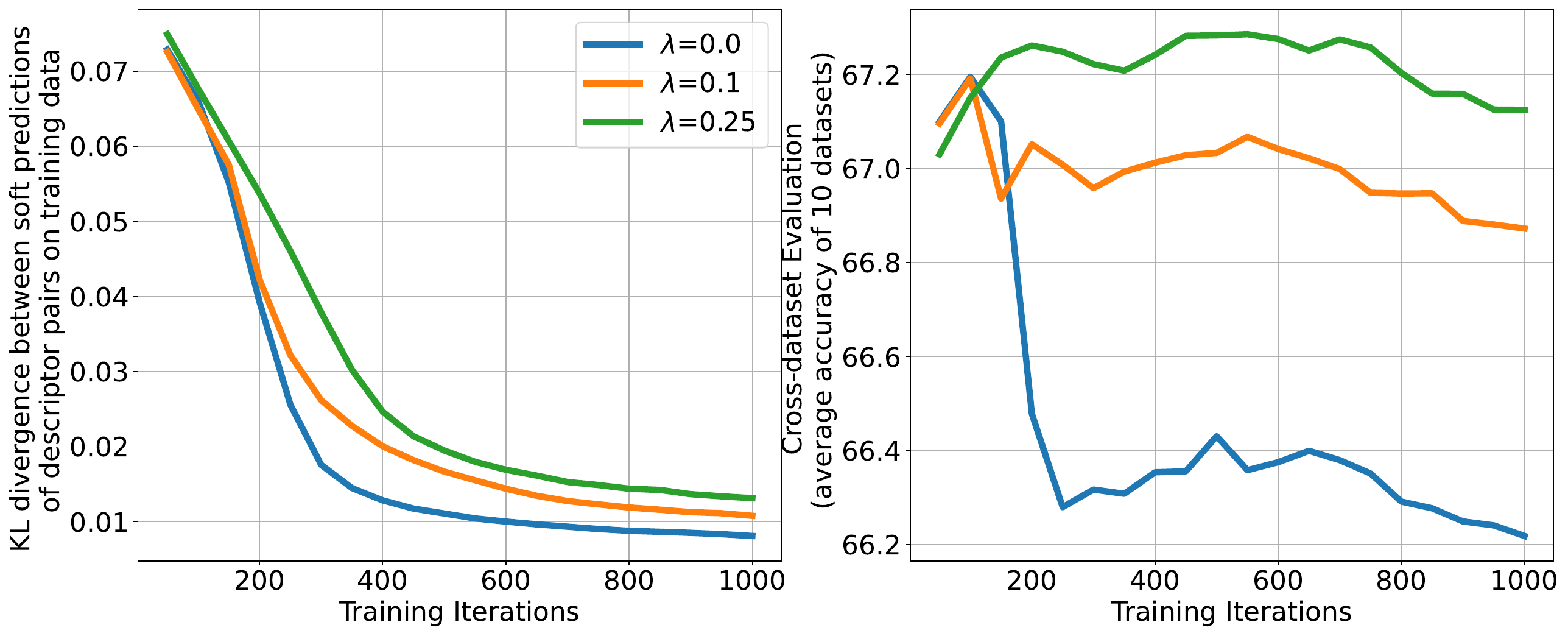}
    \caption{Varying $\lambda$ in the diversity loss. $\lambda=0$ corresponds to the standard CE loss. The left plot displays the average KL divergence between predicted class probabilities of word soup descriptors over the course of training. The right plot displays the cross-dataset accuracy for the same training runs. 
    We observe that a larger $\lambda$ leads to higher diversity among descriptors; this results in a higher test accuracy. \vspace{-1.0em} }
    \label{fig:diversity_loss}
\end{figure}

\subsection{Diversity loss}
Word soup already achieves competitive performance on most benchmarks. A reasonable next step would be to finetune using the word soup descriptors as an initialization. A variety of methods exist for few-shot finetuning of CLIP, e.g. CoOp, Clipood, and MaPLe. However, in many cases we actually see a slight decline in target accuracy after finetuning in Tab. \ref{tab:coop} ($\lambda=0$). This is because finetuning all descriptors on the same few-shot data forces text-prototypes to converge to the same locations in the embedding space, eliminating the initial diversity. Given \emph{fixed} word soup descriptors $\D^* = \{d^*_1, ..., d^*_m\}$, our training loss is:
\begin{equation}
    \ell_{\text{train}} = \EE_{ d^*_i \sim \D^* } \left[ \text{CE}(\hat{y}_{d^*_i}, (1-\lambda) y_{\text{truth}} + \lambda \hat{y}_{d^*_i, 0}) \right]
    \label{eq:diversity}
\end{equation}
where CE denotes the cross entropy loss, $\hat{y}_{d^*_i} \in \Delta_c$ ($c$ is the number of classes) denotes the soft prediction of the model with descriptor $d^*_i$; $y_{\text{truth}}$ denotes the one-hot encoding of the true label; and  $\hat{y}_{d^*_i, 0} \in \Delta_c$ denotes the soft prediction of the initial model with descriptor $d^*_i$. $\lambda \in [0,1]$ is a hyperparameter controlling the amount of regularization. $\hat{y}_{d^*_i}$ is the quantity being optimized. $\hat{y}_{d^*_i, 0}$ is the output of a softmax with temperature $\tau_0$ 
(the teacher temperature). 
As in classical knowledge distillation, it is often useful to set the teacher temperature to be different than the training temperature. 
Training the expectation directly in Eq. \ref{eq:diversity} requires storing $mc$ forward and backward passes of the text encoder in memory, which is not scalable. In practice, we use one descriptor per mini-batch and rotate among the $m$ descriptors in a round-robin fashion, but we train for the \emph{same number of iterations} as finetuning with one descriptor. 

Our training loss biases the model prediction toward the initial prediction of the model using each description, thereby maintaining the diversity of predictions present at initialization. Fig. \ref{fig:diversity_loss} verifies this interpretation by showing that training with $\lambda=0.25$ results in a higher average KL divergence between descriptor predictions $\hat{y}_{d^*_i}$ and a higher average target accuracy than training with lower $\lambda$s. 
Additionally, Tab. \ref{tab:coop} displays results for a naive CoOp ensemble and CoOp trained with regularization towards the initialization. These results show that our diversity loss results cannot be obtained by simply ensembling or regularizing predictions towards the initialization as in \citep{cha2022domain, zhu2023prompt}. 
The training does not take longer than standard cross entropy training, since only one model is trained for all descriptors. Descriptor tokens are fixed. 

\setlength\tabcolsep{4.5 pt}
\begin{table}[t!]
\scriptsize
\centering
\begin{tabular}{l c c >{\columncolor[gray]{0.9}} c >{\columncolor[gray]{0.9}}c}
\toprule

     & $m$ & Source & \textbf{XD Mean} & \textbf{DG Mean} \\
     & & INet & (10 datasets) & (4 datasets) \\

    \midrule
    
     CLIP ZS \citep{radford2021learning} &      1 &  67.1 &  65.02 & 57.22 \\

     \textcolor{gray}{CoOp \citep{zhou2022coop}}$\dagger$ & & \textcolor{gray}{71.5} &  \textcolor{gray}{63.88} & \textcolor{gray}{59.3} \\
    \textcolor{gray}{Co-CoOp \citep{zhou2022conditional}}$\dagger$ & & \textcolor{gray}{71.0} &\textcolor{gray}{65.74} & \textcolor{gray}{59.9} \\
    \textcolor{gray}{MaPLe \citep{khattak2023maple}}$\dagger$ & & \textcolor{gray}{70.7} & \textcolor{gray}{66.30} & \textcolor{gray}{60.3} \\
    \textcolor{gray}{CLIPood \citep{shu2023clipood}}$\dagger$ & & \textcolor{gray}{71.6} & & \textcolor{gray}{60.5} \\

    \midrule
         Cross Entropy (CE) &      1 &  \textbf{72.3}  & 66.80 & 60.39 \\
     {  + GPT score mean \citep{menon2022visual}} & 5.8 &  71.7 & 66.86 &  59.92 \\
     {  + Random descriptors} &      32 &  71.6 & 66.89 & 60.69 \\
     {  + Waffle CLIP \citep{roth2023waffling}} &      32 &  71.6 & 66.58 & 60.65 \\
     {  + Descriptor soup (ours)} &      16.7 &  72.1 & 67.10 & 60.70 \\
     {   $\quad$  + offset trick (ours)} &      100 &  72.1  & \textbf{67.51} & 61.01 \\
     {  + Word soup centroids (ours)} &      8 &  71.8 & 67.16 & 61.22 \\
     {  + Word soup score mean (ours)} &      8 &  71.7 & 67.43 & \textbf{61.32} \\

     {  + \textcolor{gray}{Descriptor soup upper bound}} & 11 & \textcolor{gray}{71.7} & \textcolor{gray}{67.62} & \textcolor{gray}{61.01} \\

     \midrule

     ProGrad \cite{zhu2023prompt} &      1 &  69.8  & 66.48 & 58.96 \\
     KgCoOp \cite{kan2023knowledge} &      1 &  69.2 & 66.16 & 58.64 \\
     ProDA \cite{lu2022prompt} &      32 &  70.0 & 66.23 & 58.83 \\
     Vanilla CoOp \citep{zhou2022coop} &      1 &  70.0 & 66.52 & 59.25 \\
     {  + Word soup score mean (ours)} &      8 &  \textbf{70.2} & \textbf{67.30} & \textbf{60.25}  \\
     
     \midrule

     Vanilla MaPLe \citep{khattak2023maple} &      1 &  70.7 & 66.44  & 59.32 \\
     {  + Word soup score mean (ours)} &      8 &  \textbf{70.8}  & \textbf{66.65}  & \textbf{60.20} \\

     \midrule
    
     Vanilla CLIPood \citep{shu2023clipood} &      1 &  \textbf{72.9} & 66.50 & 60.47 \\
     {  + Word soup score mean (ours)} &      8 &  72.0 & \textbf{67.42}  & \textbf{61.23} \\
     
\bottomrule
  \end{tabular}
  \caption{Comparison with few-shot methods and few-shot methods stacked with ZS methods. $\dagger$ indicates author-reported numbers on the same datasets with the same train-test splits. Other numbers are our reproductions. 
  All methods except the upper bound were trained on 3 random 16-shot splits of ImageNet. 
  $m$ indicates number of descriptors used. 
  Either our descriptor soup with the offset trick or our word soup achieves the best accuracy on average. 
  We use the ViT/B-16 CLIP model. Detailed results see Tab. \ref{tab:ft-appendix} in the Appendix. \vspace{-1.0em}
  }
  \label{tab:ft}
\end{table}
\setlength\tabcolsep{6 pt}


\section{Results}
\begin{figure}[t!]
    \centering
    \includegraphics[width=1.\linewidth]{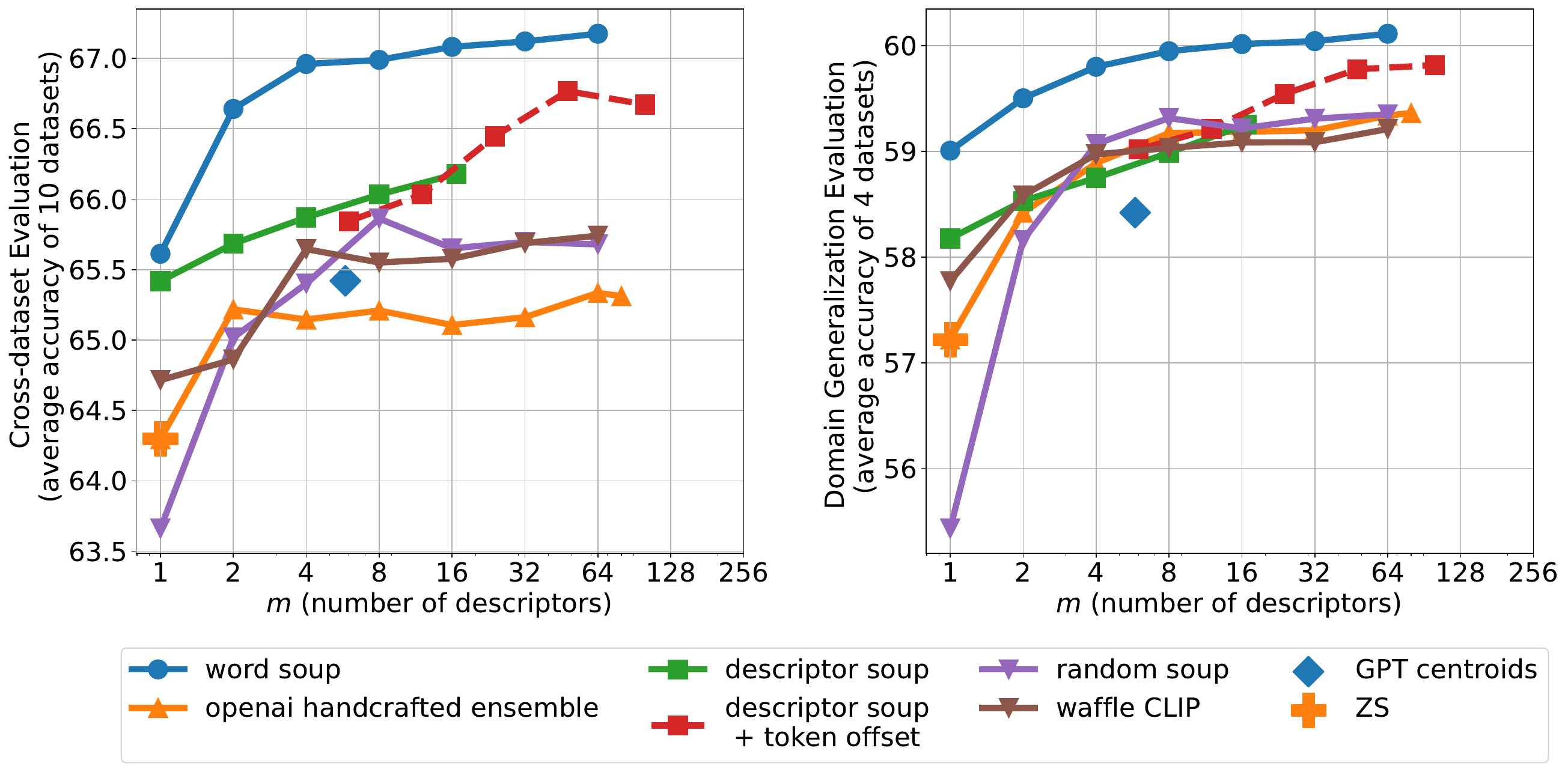}
    \caption{Comparison of our soups with ZS baselines for varying $m$ on XD and DG evaluations. 
    This experiment uses the same settings as Tab. \ref{tab:pretrained}.
    Our word soup achieves the best accuracies for all $m$. This shows that word soup is more descriptor efficient than baseline ZS methods.
    }
    \label{fig:m_ablations_pretrained}
\end{figure}

We present the main few-shot results in Tab. \ref{tab:ft}. 
The goal this section is to demonstrate the following in the OOD setting:
\begin{enumerate}
    \item \textbf{Complementary to existing few-shot methods: } Stacking either descriptor soup or word soup on top of traditional finetuning baselines (Cross Entropy, MaPLe, Clipood, or CoOp) improves target accuracy, exceeding current published state-of-art. (Tab. \ref{tab:ft})
    \item \textbf{Parameter Efficiency: } Our method is more parameter efficient than CoOp due to the discrete nature of word soup tokens. We additionally compare to other PEFT methods: VPT \cite{jia2022visual}, bitfi t\cite{zaken2021bitfit}, CLIP-adapter \cite{gao2023clip}, SSF \cite{lian2022scaling}, LoRA \cite{hu2021lora}, and adapter \cite{houlsby2019parameter}. (Fig. \ref{fig:parameter_efficiency})
    \item \textbf{Descriptor Efficiency: } We outperform prior state-of-the-art ZS methods with only 1 or 2 descriptors. Therefore, unlike some prior methods, our method is \emph{not} primarily driven by ensembling. (Fig. \ref{fig:m_ablations_pretrained})
\end{enumerate}

 \vspace{-0.8em}
\paragraph{Datasets} We train on random 16-shot splits of ImageNet-1K \citep{ILSVRC15} and test on 14 unseen target datasets: Caltech-101 \citep{li_andreeto_ranzato_perona_2022}, Oxford-Pets \citep{parkhi2012cats}, Stanford-Cars \citep{krause20133d}, Flowers-102 \citep{nilsback2008flowers102}, Food-101 \citep{bossard2014food101}, FGVC-Aircraft \citep{maji2013fgvcaircraft}, SUN-397 \citep{xiao2010sun397}, Describable-Textures (DTD) \citep{cimpoi2013dtd}, EuroSAT \citep{helber2019eurosat}, UCF-101 (an action recognition dataset) \citep{soomro2012ucf101}, ImageNet-V2 \citep{recht2019imagenetv2}, ImageNet-Sketch \citep{wang2019imagenetsketch}, ImageNet-A (natural adversarial examples) \citep{hendrycks2021natural}, and ImageNet-R \citep{hendrycks2021many}. The last four datasets are domain-shifted versions of ImageNet containing images from the ImageNet-1K label space. 

 \vspace{-0.8em}
\paragraph{Experimental Setting} 
All baselines and methods are trained on 16-shot ImageNet-1K data
and tested on the indicated target datasets. \emph{Hyperparameters:} We tune parameters on a withheld validation set. Word soup (Alg. 2) has three parameters: $k_0$, $k_1$ and patience. The diversity loss has two parameters: $\lambda$ and $\tau_0$. These 5 parameters are constant across all experiments. We tune the learning rate separately for each baseline, but keep all other training parameters consistent across methods. 
We report temperature, batch size, optimizer, EMA setting, token length, initialization and other training details in Appendix A.
We discuss the difference between centroid and score mean evaluation in Appendix D.

 \vspace{-0.8em}
\paragraph{Discussion}
In Tab. \ref{tab:ft}, we first observe that stacking our word soup method on top of CE, CoOp, MaPLe, or CLIPood achieves approximately 0.8-1.0\% increase in average target accuracy for both XD and DG benchmarks. Due to the space limitation, we only compare word soup with other ZS methods when combined with CE, since CE achieves the highest XD accuracy out of the 4 finetuning methods. $m$ indicates the number of descriptors for each label, on average. 
The greedy descriptor soup can be augmented using our token offset trick, which uses 6 augmented copies of each descriptor. The token offset trick improves accuracy by 0.4\% and 0.3\% on XD and DG, resp. but at a significant computational cost. 
The greedy word soup matches the performance of the augmented descriptor soup without the additional computational cost. 
Overall, the best OOD accuracy is achieved by either the descriptor soup with token offsets or word soup. 

 \vspace{-0.8em}
\paragraph{Ablation Study}
An ablation study on our soup methods with varying $m$ is presented in Fig. \ref{fig:m_ablations_pretrained}. On both benchmarks, our word soup performs best for all $m$. 
We note that the word soup with $m=2$ already outperforms all ZS baselines for all values of $m$ up to 64. This result indicates that, unlike state-of-the-art ZS methods, ensembling is not the main ingredient of our method. Additional ablation studies are presented in Appendix E.

 \vspace{-0.8em}
 \paragraph {Parameter Efficiency and Computational Efficiency} A discussion regarding efficiency of our methods is deferred to Appendix E.

\section{Conclusion}
In this paper, we proposed descriptor and word soups to tackle the cross-dataset and domain generalization problems. Descriptor soup greedily selects a set of descriptors by maximizing training accuracy on a source dataset. Word soup builds a chain of words using a similar greedy procedure. These greedy soup methods achieve higher target classification accuracy than prior descriptor-based methods by explicitly maximizing training accuracy. 
We further proposed a loss function to preserve word soup diversity throughout finetuning. 
When using word soup for initialization and finetuning with the diversity loss, we can significantly improve the accuracy of existing few-shot OOD finetuning methods. Compared to all baselines, word soup achieves the best trade-off between parameter efficiency and target accuracy.

\section*{Acknowledgements}

DISTRIBUTION STATEMENT A. Approved for public release. Distribution is unlimited.

This material is based upon work supported by the Under Secretary of Defense for Research and Engineering under Air Force Contract No. FA8702-15-D-0001. Any opinions, findings, conclusions or recommendations expressed in this material are those of the author(s) and do not necessarily reflect the views of the Under Secretary of Defense for Research and Engineering.

{
    \small
    \bibliographystyle{ieeenat_fullname}
    \bibliography{main}
}

\clearpage
\setcounter{page}{1}
\maketitlesupplementary
\appendix

\begin{algorithm}[tb]
   \caption{Descriptor soup pseudo-code, PyTorch-like}
   \label{alg:descriptor_soup}
   \begin{minted}[fontsize=\scriptsize]{python}
# '@' means matrix multiplication in Python.
# '+' means concatenation when operating on lists.
# Inputs: L2-normalized image_embeddings, y_truth
# classnames: list of classnames in English
# model: CLIP-style model
# descriptions: list of descriptions from an LLM
# Hyperparameters: m (number of members in the soup)
descriptions = ['which has legs.', 'which can swim.', ... ]

def get_accuracy(image_embeddings, text_embeddings, 
                 y_truth):
    scores = image_embeddings @ text_embeddings.T
    return (scores.argmax(dim=1) == y_truth).mean()

def get_description_embeddings(description):
    d = tokenizer(['a photo of ' + classname + ', '  +
        description for classname in classnames])
    return normalize(model.encode_text(d))
    
accuracies = []
for description in descriptions:
    text_embeddings = get_description_embeddings(
                          description)
    accuracies.append(get_accuracy(
        image_embeddings, text_embeddings, y_truth))
# sort descriptions by accuracies
descriptions_sorted = descriptions[
    accuracies.sort(descending=True).indices]

# initialize with best descriptor
soup, accuracy = [descriptions_sorted[0]], accuracies[0]

# greedy selection
for description in descriptions_sorted:
    soup_embeddings = stack(
        [get_description_embeddings(description) 
            for description in soup + [description] ] )
    text_embeddings = normalize(
        soup_embeddings.mean(dim=0))
    if get_accuracy(image_embeddings, 
        text_embeddings, y_truth) > current_acc:
        soup = soup + [description]
        
return soup[:m]
    \end{minted}
\end{algorithm}

\begin{algorithm}[tb]
   \caption{Word soup pseudo-code, PyTorch-like}
   \label{alg:word_soup}
   \begin{minted}[fontsize=\scriptsize]{python}
# Hyperparameters: k0, k1, m, patience
# Inputs: L2-normalized image_embeddings, y_truth

words = ["the", "of", "and", ... ]
accuracies = []
for word in words:
    text_embeddings = get_description_embeddings(word)
    accuracies.append(get_accuracy(
        image_embeddings, text_embeddings, y_truth))
# sort descriptions by accuracies
words = words[accuracies.sort(descending=True).indices]

soup = []
for repeat m times:
    first_word = random.shuffled(words[0:k0])[0]
    word_chain = first_word
    accuracy = get_accuracy(image_embeddings, 
        get_description_embeddings(word_chain), y_truth)
    words_k1 = random.shuffled(words[0:k1])[0:patience]
    
    # greedy selection
    for word in words_k1:
        text_embeddings = get_description_embeddings(
            word_chain + " " + word) 
        next_accuracy = get_accuracy(image_embeddings, 
            text_embeddings, y_truth)
        if next_accuracy > accuracy:
            word_chain = word_chain + " " + word
    soup = soup + [word_chain]
        
return soup
    \end{minted}
\end{algorithm}


\section*{Limitations}
Similar to many related works, the main limitation of our work is that we require the source dataset to cover a broad range of classes (e.g. ImageNet). As a counter example, we cannot hope to train on pets classification and generalize to ImageNet. We highlighted this limitation in Table \ref{tab:descriptor_examples} of the main paper (top) with qualitative examples.

\section{Training details}
Images are not augmented during the greedy descriptor selection process; image augmentation during finetuning is consistent with prior work. Descriptors are always selected using the pretrained model parameters. Selecting descriptors based on finetuned model weights would be sub-optimal, since the pretrained text encoder captures a richer set of textual information. Remaining details are organized in Table \ref{tab:training-details}. Mini-batches are randomly sampled, but with exactly one sample per label per batch. Cross entropy and CLIPood both tune the last three layers of the image and text encoders, in addition to a shallow text prompt (like CoOp) at a higher learning rate. The only difference between Cross entropy and CLIPood is the loss function; the latter method uses an adaptive margin. We use cross entropy loss for all baselines except ProDA and ProGrad. ProDA and ProGrad consume more GPU memory during training, so we were unable to fit them onto a single A40 GPU when training with cross entropy. Consequently, we were forced to use a CLIP-like contrastive loss for these two methods to reduce the number of text encoder evaluations.

\setlength\tabcolsep{3.1 pt}
\begin{table}[t!]
\centering
\begin{tabular}{l r }
\toprule
\multicolumn{2}{c}{\textbf{General Parameters}} \\
\midrule
batch size & 64 \\
learning rate & tuned per method \\
weight decay & 1e-5 \\
number of iterations & 750 \\
learning rate decay & none \\
softmax temperature & 60 \\
optimizer & SGD momentum=0.9  \\
label smoothing & 0 \\
EMA weight averaging $\beta$ & 0.995 \\
\bottomrule
\toprule
\multicolumn{2}{c}{\textbf{Prompt Tuning Parameters}} \\
\midrule
CoOp prompt length & 3 \\
CoOp prompt depth & 1 (shallow) \\
MaPLe prompt depth & 3 \\
MaPLe prompt length & 3 \\
CoOp prompt initialization & ``a photo of'' \\
text prompt learning rate multiplier & 10 $\times$ \\
\bottomrule
\toprule
\multicolumn{2}{c}{\textbf{Word Soup and Diversity Loss Parameters}} \\
\midrule
$k_0$ & 250 \\
$k_1$ & 1000 \\
patience & 250 \\
$\lambda$ &  0.25 \\
$\tau_0$ & 10 \\

\bottomrule
\toprule
\multicolumn{2}{c}{\textbf{Optimal Learning Rates}} \\
\midrule
Cross entropy & 2e-5 \\
CLIPood \citep{shu2023clipood} & 2e-5 \\
CoOp \citep{zhou2022coop} & 8e-5 \\
MaPLe \citep{khattak2023maple} & 0.025 \\
KgCoOp \cite{kan2023knowledge} &  4e-5 \\
ProDA \cite{lu2022prompt} &  3.2e-4 \\
ProGrad \cite{zhu2023prompt}  &  1.28e-3 \\
VPT \cite{jia2022visual} &  0.8 \\
bitfit \cite{zaken2021bitfit} &  1.25e-4 \\
CLIP-adapter \cite{gao2023clip} &  6e-3 \\
SSF \cite{lian2022scaling} &  1e-4 \\
adapter \cite{houlsby2019parameter} &  2.5e-3 \\
LoRA \cite{hu2021lora} &  1e-5 \\
\bottomrule

  \end{tabular}
  \caption{Miscellaneous training details for training on 16-shot ImageNet-1K in the OOD setting. }
  \label{tab:training-details}
\end{table}


\section{Additional Word Soup Motivation} 
\begin{figure}[t!]
    \centering
    \includegraphics[width=1.\linewidth]{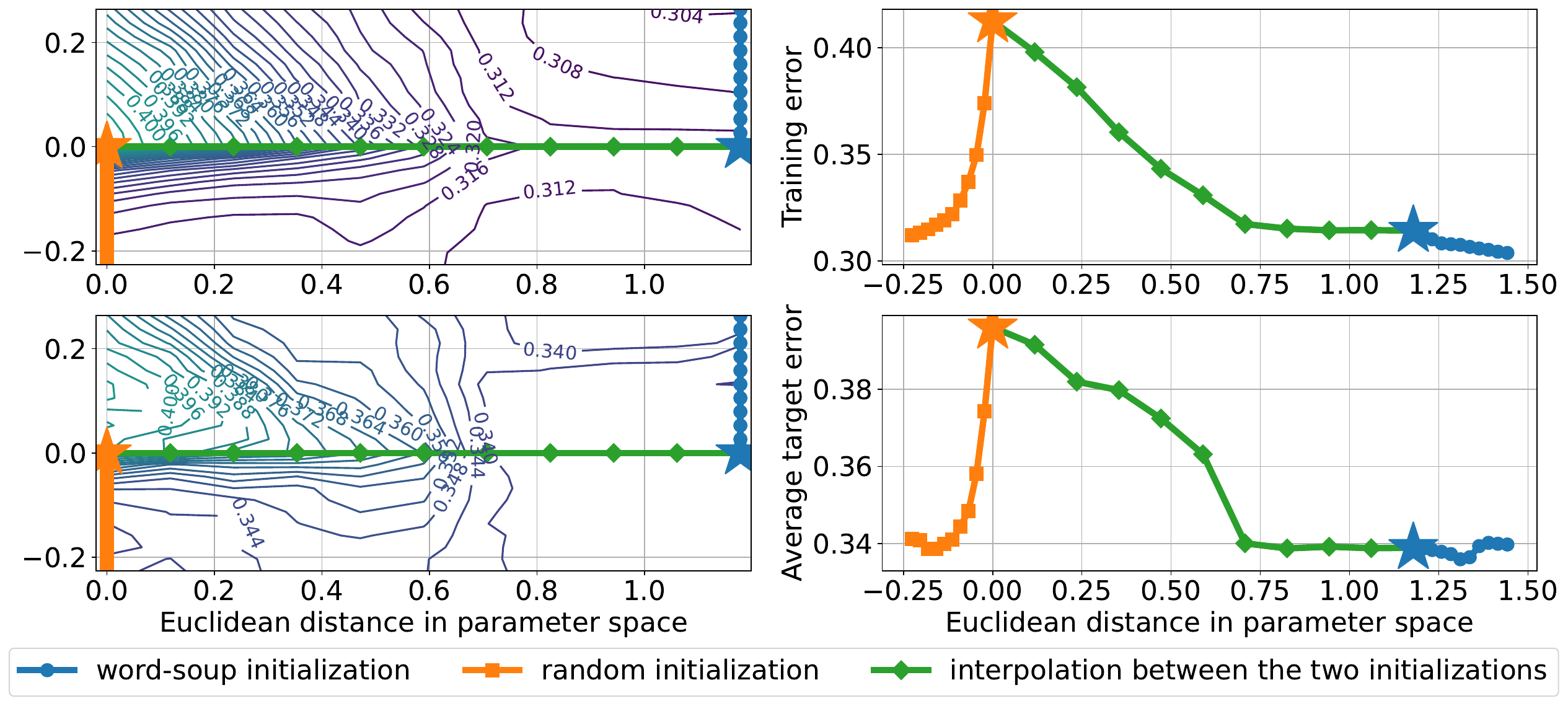}
    \caption{Contour plot of the 0-1 loss over the 2D parameter space spanned by two initializations (indicated by stars) and the fine-tuned parameters. The orange and blue stars indicate the random initialization and word soup initialization, resp. The top and bottom rows plot the 0-1 loss on the training and test data (average of 10 test datasets), resp. For this figure, we train 10 descriptor tokens. The plots on the right indicate the loss value at the corresponding locations in the contour plots on the left, for better visualization. We observe that the word soup initialization lies in a lower and flatter region, compared to the random initialization. Consequently, finetuning from the word soup initialization results in lower training and test errors compared to finetuning from the random initialization.}
    \label{fig:loss_landscape}
\end{figure}

A natural baseline for word soup is soft prompt tuning (CoOp), since the former method can be thought of as ``discrete'' prompt tuning. Soft prompt tuning optimizes over a continuous parameter space using gradient descent, whereas word soup optimizes over a discrete parameter space using a greedy algorithm. Many prior works (e.g. \citep{wortsman2022model, wortsman2022robust}) observe that gradient descent is limited to a narrow convex basin around the initialization, when finetuning a pretrained deep model. This can be shown by linearly interpolating between the pretrained and finetuned parameters, similar to Fig. \ref{fig:loss_landscape}. In this figure, we plot in orange both the source and target error for interpolations between a \emph{randomly initialized} descriptor (orange star) and the finetuned soft descriptor. The resulting soft descriptor lies at the bottom of a sharp loss basin. On the other hand, the \emph{word soup initialized} descriptor (blue star) lies at an equally low but much flatter region of the loss landscape. Finetuning from this initialization leads to a lower error on both source and target data, as indicated in blue. This visualization suggests that our word soup algorithm finds robust flat minima, since it is not limited to a narrow loss basin like gradient descent methods.

\setlength\tabcolsep{3.5 pt}
\begin{table*}[th!]
\small
\centering
\begin{tabular}{l ccc ccc ccccc >{\columncolor[gray]{0.9}}ccccc >{\columncolor[gray]{0.9}}c}
\toprule
& \textbf{Source} & \multicolumn{11}{c}{\textbf{Cross-dataset Evaluation Targets}} & \multicolumn{5}{c}{\textbf{Domain Generalization Targets}} \\

\cmidrule(lr){2-2} \cmidrule(lr){3-13} \cmidrule(lr){14-18}

     & \rotatebox{90}{ INet } & \rotatebox{90}{ Caltech } & \rotatebox{90}{ Pets } & \rotatebox{90}{ Cars } & \rotatebox{90}{ Flowers } & \rotatebox{90}{ Food } & \rotatebox{90}{ Aircraft } & \rotatebox{90}{ SUN } & \rotatebox{90}{ DTD } & \rotatebox{90}{ EuroSAT } & \rotatebox{90}{ UCF } & \rotatebox{90}{ \textbf{Mean} } & \rotatebox{90}{ INet-V2 } & \rotatebox{90}{ Sketch } & \rotatebox{90}{ INet-A } & \rotatebox{90}{ INet-R } & \rotatebox{90}{ \textbf{Mean} }\\
     
    \midrule
    
     CLIP ZS &  67.1 & 93.3 & 89.0 & 65.4 & 71.0 & 85.7 & 25.0 & 63.2 & 43.6 & 46.7 & 67.4 & 65.02 & 61.0 & 46.6 & 47.2 & 74.1 & 57.22 \\

     \textbf{Word soup} & 68.8& 94.1& 89.5& 65.9& \textbf{72.6} & \textbf{86.3} & \textbf{26.1} & 67.2& 45.3& \textbf{53.9} & 67.8& \textbf{66.87} & 62.6& 49.0& \textbf{50.4} & 77.0& 59.73 \\
     
    Vanilla CoOp  & 68.7& 94.4& 90.2& \textbf{66.1} & 70.9 & 85.8& 26.0& 66.7& \textbf{47.4} & 50.1 & 68.9& 66.63& 61.9& 48.6& 49.8& 76.7& 59.26 \\
    
    \textbf{ $\quad$ + Word soup} &    \textbf{69.1} &  \textbf{94.6} & \textbf{91.1} & 65.2& 71.8& 86.0& 25.1& \textbf{67.4} & 46.0& 51.9& \textbf{69.1} & 66.82& \textbf{62.7} & \textbf{49.4} & 50.3& \textbf{78.0} & \textbf{60.09} \\

\bottomrule
  \end{tabular}
  \caption{ 
  Experiments using a different source dataset (a 16-shot subset of LAION-2B queried using ImageNet label names). Settings are identical to Table \ref{tab:ft-appendix} (the expanded form of Table \ref{tab:ft} in the main paper). }
  \label{tab:laion-source}
\end{table*}
\setlength\tabcolsep{6 pt}

\section{Token offset trick (for Descriptor Soup)}
We propose a novel trick to augment/diversify the descriptors at test time to further increase the target accuracy of descriptor soups. This trick does not improve the performance of word soups significantly. Unlike the vision encoder, which has a cls token at a fixed position (either prepended or appended to the image tokens), the CLIP text encoder does not have a separate cls token. Instead, CLIP uses the output embedding which corresponds to the position of the end-of-sentence token in the input. In classification problems, the text inputs are generally short compared to the context size (number of total tokens). Consequently, the end-of-sentence token is always near the beginning of the sequence, with the remainder padded by null tokens. In this regime, there is never any information at the end of the input token sequence to attend to, so a large portion of the information in the pretrained model is not used. We remedy this inefficient use of pretrained parameters by shifting the description toward the end of the sequence by $t$ tokens. For example, if $t=5$, we have:
\begin{itemize}
    \item \textbf{original: } a photo of a dog, which may be large or small.
    \item \textbf{augmented: } a photo of a dog, ! ! ! ! ! which may be large or small. (``!'' denotes the null token)
\end{itemize}
For all experiments with token offsets, we set $t=\{0, 5, 10, 15, 20, 25\}$ for a total of 6 augmented copies per descriptor. This diversifies the text embeddings at the expense of increasing the text centroid evaluation time 6-folds.

\section{Centroid vs. Score Mean Evaluation}
In this work, we presented both centroid and score mean results for both our soup methods and ensemble baselines. Centroid evaluation refers to averaging the text features among descriptors before calculating the cosine similarity between image and text features. Score mean evaluation refers to calculating the cosine similarity between image and text features and then averaging the similarity scores among descriptors. 

Concretely, let there be $m$ descriptors and $c$ classes. Let $\mathbf{x}_{I}$ denote a normalized image feature and $\mathbf{x}_{T, k}^j$ denote the normalized text feature corresponding to class $k$ and descriptor $j$; $k \in [1:c]$ and $j \in [1:m]$.

The predicted score for class $k$ using centroid evaluation, $s_k$, is defined as:
\begin{equation*}
    \begin{split}
        \overline{\mathbf{x}}_{T, k} = \frac{1}{m} \sum_{j=1}^{m} \mathbf{x}_{T, k}^j
    \end{split}
\end{equation*}
$$ s_k = \left\langle \mathbf{x}_{I} , \frac{\overline{\mathbf{x}}_{T, k}} {\| \overline{\mathbf{x}}_{T, k}\|} \right\rangle $$

The predicted score for class $k$ using score mean evaluation is defined as:
\begin{equation*}
    s_k =  \frac{1}{m} \sum_{j=1}^{m}  \left\langle \mathbf{x}_{I} ,  \mathbf{x}_{T, k}^j \right\rangle
\end{equation*}

Empirically, we found that score mean evaluation usually leads to small numerical improvements. However, in large scale applications where retrieval speed is crucial, centroid evaluation can be more efficiently implemented than score mean evaluation, due to the existence of fast nearest neighbor retrieval frameworks.

\begin{figure}[t]
  \centering
  \includegraphics[width=1.0\linewidth]{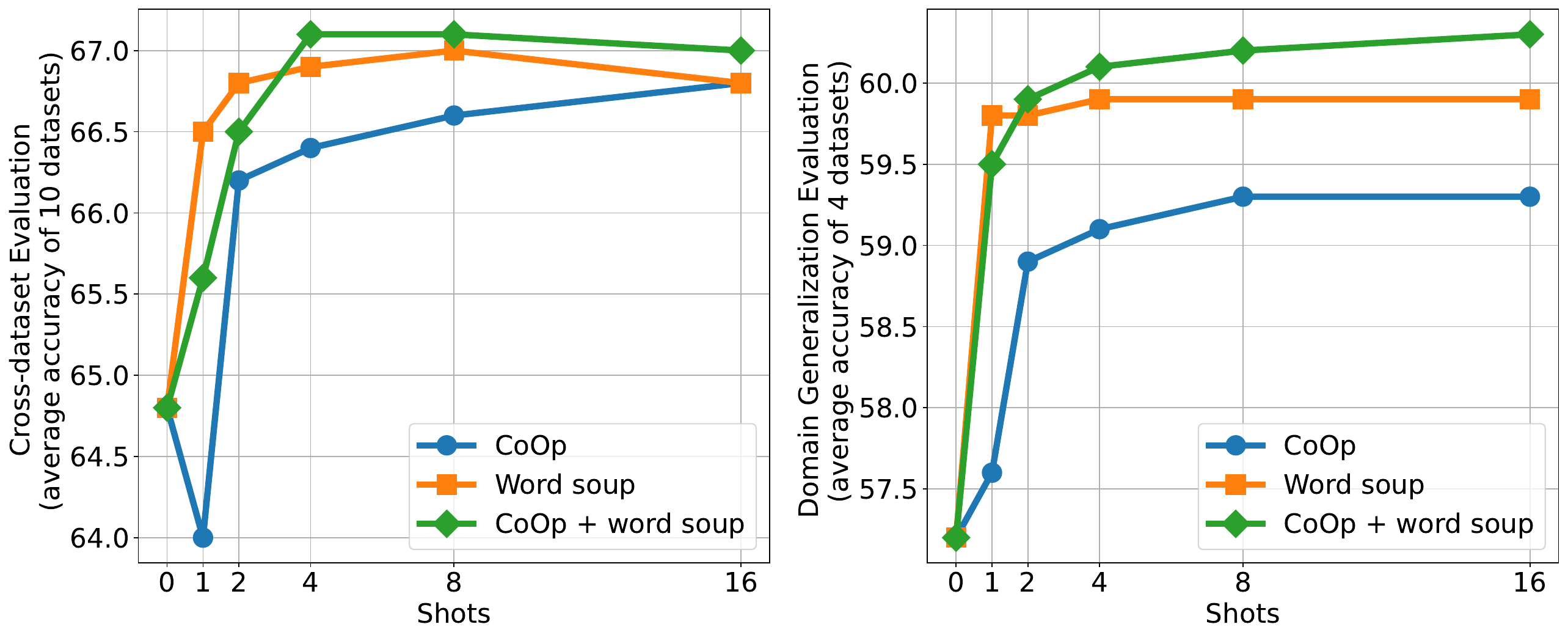}
   \caption{Different number of shots. We experiment with the same 14 datasets as the main paper and report average of 3 random trials. We report average target accuracies over 10 diverse datasets (left) and 4 ImageNet shifts (right). Here we verify that the improvements of both word soup and CoOp + word soup over CoOp are resilient to the number of shots. Indeed, we emphasize that \textbf{word soup is very resilient in extreme low shot scenarios due to the low number of parameters.}}
   \label{fig:shots-rebuttal}
\end{figure}

\section{Additional Ablation Studies}
We present additional ablation studies in Table \ref{tab:laion-source} and Figure \ref{fig:shots-rebuttal}. Table \ref{tab:laion-source} presents OOD generalization results with a different source data set. Figure \ref{fig:shots-rebuttal} presents results with different number of shots.

\paragraph{Parameter Efficiency}
Fig. \ref{fig:parameter_efficiency} compares the parameter efficiency of our word soups against PEFT baselines. 
We observe that word soup can achieve the maximal CoOp accuracy using 25$\times$ and 70 $\times$ fewer parameters on the XD and DG benchmarks, resp. This impressive reduction in parameter storage requirements is due to the discrete nature of word soup parameters. A discrete token requires only one integer parameter, while a soft token requires 512 floating-point parameters. 

 \vspace{-0.8em}
\paragraph{Computational Efficiency}
We emphasize that our method adds negligible test time computation, despite requiring $m$ text encoder evaluations per label. 
For classification tasks, more time is spent processing image data compared to text data. 
For example, the evaluation of the $m=8$ word soup in Table \ref{tab:ft} took 239 seconds, of which 234 seconds were spent evaluating image embeddings and only 4.6 seconds were spent evaluating text embeddings. 

\setlength\tabcolsep{3.95 pt}
\begin{table*}[t!]
\scriptsize
\centering
\begin{tabular}{l c ccc ccc ccccc >{\columncolor[gray]{0.9}}ccccc >{\columncolor[gray]{0.9}}c}
\toprule
&& \textbf{Source} & \multicolumn{11}{c}{\textbf{Cross-dataset Evaluation Targets}} & \multicolumn{5}{c}{\textbf{Domain Generalization Targets}} \\

\cmidrule(lr){3-3} \cmidrule(lr){4-14} \cmidrule(lr){15-19}

     & $m$ & \rotatebox{90}{ INet } & \rotatebox{90}{ Caltech } & \rotatebox{90}{ Pets } & \rotatebox{90}{ Cars } & \rotatebox{90}{ Flowers } & \rotatebox{90}{ Food } & \rotatebox{90}{ Aircraft } & \rotatebox{90}{ SUN } & \rotatebox{90}{ DTD } & \rotatebox{90}{ EuroSAT } & \rotatebox{90}{ UCF } & \rotatebox{90}{ \textbf{Mean} } & \rotatebox{90}{ INet-V2 } & \rotatebox{90}{ Sketch } & \rotatebox{90}{ INet-A } & \rotatebox{90}{ INet-R } & \rotatebox{90}{ \textbf{Mean} }\\
     
    \midrule

    CLIP ZS &      1 &  67.1 & 93.3 & 89.0 & 65.4 & 71.0 & 85.7 & 25.0 & 63.2 & 43.6 & 46.7 & 67.4 & 65.02 & 61.0 & 46.6 & 47.2 & 74.1 & 57.22 \\
     Vanilla CoOp &      1 &  70.0 & 94.6 & 91.2 & 65.4 & 71.2 & 86.3 & 24.6 & 66.9 & \textbf{48.0} & 48.3 & 68.7 & 66.52 & 63.2 & 48.4 & 49.2 & 76.2 & 59.25 \\
     {  + word soup} &      8 &  69.6 & 94.6 & 90.8 & 65.2 & 70.3 & 86.0 & 24.8 & 66.9 & 47.6 & 50.7 & 69.0 & 66.59 & 62.9 & 48.2 & 49.6 & 76.3 & 59.26 \\
     CoOp ensemble &      8 &  69.8 & 94.4 & \textbf{91.5} & \textbf{66.2} & \textbf{72.6} & \textbf{86.6} & \textbf{25.7} & 67.7 & 46.4 & 47.9 & 67.8 & 66.68 & 63.0 & 48.4 & 49.6 & 75.8 & 59.18 \\
     \midrule
     
     CoOp regularized towards initialization & 1 &  70.2 & \textbf{94.8} & 91.1 & 65.4 & 72.1 & 86.2 & 24.8 & 67.6 & 46.2 & 52.7 & 69.0 & 66.97 & \textbf{63.6} & 49.1 & 49.6 & 77.5 & 59.94 \\
     { $\quad$ + word soup} &      8 &  69.9 & 94.7 & 90.1 & 64.7 & 71.8 & 85.5 & 25.0 & 67.4 & 45.5 & 53.6 & 68.7 & 66.69 & 63.4 & 49.2 & 49.9 & 77.7 & 60.05 \\
     CoOp with label smoothing &      1 &  70.1 & 94.5 & 90.6 & 64.9 & 72.0 & 85.8 & 24.6 & 67.3 & 45.4 & 50.0 & 68.6 & 66.37 & 63.4 & 49.1 & \textbf{50.2} & 77.6 & 60.09 \\
     { $\quad$ + word soup} &      8 &  69.9 & 94.5 & 89.9 & 64.9 & 71.7 & 85.2 & 25.0 & 66.8 & 44.8 & 50.0 & 68.3 & 66.13 & 63.6 & 49.3 & 50.1 & 77.7 & 60.16 \\
     CoOp + word soup ($\lambda=0$) &      8 &  69.8 & 94.3 & 90.8 & 64.8 & 71.1 & 86.0 & 24.1 & 67.2 & 46.8 & 48.4 & 68.8 & 66.21 & 63.2 & 48.3 & 49.0 & 76.1 & 59.15 \\
     {$\quad$ + \textbf{our diversity loss} ($\lambda=0.25$)} &      8 &  \textbf{70.2} & 94.7 & 91.0 & 65.4 & 72.3 & 86.0 & 24.8 & \textbf{67.8} & 45.9 & \textbf{55.2} & \textbf{69.2} & \textbf{67.23} & 63.6 & \textbf{49.3} & 50.1 & \textbf{77.9} & \textbf{60.20} \\
    
\bottomrule
  \end{tabular}
  \caption{Ablation results to support the diversity loss. ``Vanilla CoOp + word soup'' refers to naively appending the word soup descriptors trained on the pretrained model to the separately trained soft CoOp prompts. ``CoOp ensemble'' refers to ensembling $m$ randomly-initialized soft descriptors. This requires running CoOp $m$ times, but offers negligible gains in accuracy. In the second half of the table, we fix the descriptor tokens and train the prompt tokens only. We first run CoOp with standard CE training ($\lambda=0$) and observe a decrease in accuracy compared to the naive ``Vanilla CoOp + word soup'' baseline, caused by the diversity collapse issue observed in Figure \ref{fig:diversity_loss}. We then attempt to simply minimize the KL divergence between the training prediction and the initial prediction; this shows that the diversity loss is not simply a form of regularization towards the initialization as in MIRO \citep{cha2022domain} and ProGrad \citep{zhu2023prompt}. Finally, we train using our diversity loss with $\lambda=0.25$, which achieves a 1\% increase in accuracy on average. Average of 3 trials. This is an expanded version of Table \ref{tab:coop} in the main paper. }
  \label{tab:coop-appendix}
\end{table*}
\setlength\tabcolsep{6 pt}

\setlength\tabcolsep{3.1 pt}
\begin{table*}[t!]
\scriptsize
\centering
\begin{tabular}{l c ccc ccc ccccc >{\columncolor[gray]{0.9}}ccccc >{\columncolor[gray]{0.9}}c}
\toprule
&& \textbf{Source} & \multicolumn{11}{c}{\textbf{Cross-dataset Evaluation Targets}} & \multicolumn{5}{c}{\textbf{Domain Generalization Targets}} \\

\cmidrule(lr){3-3} \cmidrule(lr){4-14} \cmidrule(lr){15-19}

     & $m$ & \rotatebox{90}{ INet } & \rotatebox{90}{ Caltech } & \rotatebox{90}{ Pets } & \rotatebox{90}{ Cars } & \rotatebox{90}{ Flowers } & \rotatebox{90}{ Food } & \rotatebox{90}{ Aircraft } & \rotatebox{90}{ SUN } & \rotatebox{90}{ DTD } & \rotatebox{90}{ EuroSAT } & \rotatebox{90}{ UCF } & \rotatebox{90}{ \textbf{Mean} } & \rotatebox{90}{ INet-V2 } & \rotatebox{90}{ Sketch } & \rotatebox{90}{ INet-A } & \rotatebox{90}{ INet-R } & \rotatebox{90}{ \textbf{Mean} }\\
     
    \midrule
    
     CLIP ZS \citep{radford2021learning} &      1 &  67.1 & 93.3 & 89.0 & 65.4 & 71.0 & 85.7 & 25.0 & 63.2 & 43.6 & 46.7 & 67.4 & 65.02 & 61.0 & 46.6 & 47.2 & 74.1 & 57.22 \\

     \textcolor{gray}{CoOp \citep{zhou2022coop}}$\dagger$ & & \textcolor{gray}{71.51} & \textcolor{gray}{93.70} & \textcolor{gray}{89.14} & \textcolor{gray}{64.51} & \textcolor{gray}{68.71} & \textcolor{gray}{85.30} & \textcolor{gray}{18.47} & \textcolor{gray}{64.15} & \textcolor{gray}{41.92} & \textcolor{gray}{46.39} & \textcolor{gray}{66.55} & \textcolor{gray}{63.88} & \textcolor{gray}{64.20} & \textcolor{gray}{47.99} & \textcolor{gray}{49.71} & \textcolor{gray}{75.21} & \textcolor{gray}{59.3} \\
    \textcolor{gray}{Co-CoOp \citep{zhou2022conditional}}$\dagger$ & & \textcolor{gray}{71.02} & \textcolor{gray}{{94.43}} & \textcolor{gray}{90.14} & \textcolor{gray}{65.32} & \textcolor{gray}{71.88} & \textcolor{gray}{86.06} & \textcolor{gray}{22.94} & \textcolor{gray}{67.36} & \textcolor{gray}{45.73} & \textcolor{gray}{45.37} & \textcolor{gray}{68.21} & \textcolor{gray}{65.74} & \textcolor{gray}{64.07} & \textcolor{gray}{48.75} & \textcolor{gray}{50.63} & \textcolor{gray}{76.18} & \textcolor{gray}{59.9} \\
    \textcolor{gray}{MaPLe \citep{khattak2023maple}}$\dagger$ & & \textcolor{gray}{70.72} & \textcolor{gray}{93.53} & \textcolor{gray}{{90.49}} & \textcolor{gray}{65.57} & \textcolor{gray}{{72.23}} & \textcolor{gray}{86.20} & \textcolor{gray}{24.74} & \textcolor{gray}{67.01} & \textcolor{gray}{{46.49}} & \textcolor{gray}{48.06} & \textcolor{gray}{68.69} & \textcolor{gray}{66.30} & \textcolor{gray}{64.07} & \textcolor{gray}{49.15} & \textcolor{gray}{{50.90}} & \textcolor{gray}{76.98} & \textcolor{gray}{60.3} \\
    \textcolor{gray}{CLIPood \citep{shu2023clipood}}$\dagger$ & & \textcolor{gray}{71.6} & &&&&&&&&&&& \textcolor{gray}{64.9} & \textcolor{gray}{49.3} & \textcolor{gray}{50.4} & \textcolor{gray}{77.2} & \textcolor{gray}{60.5} \\

    \midrule
         Cross Entropy (CE) &      1 &  \textbf{72.3} & 94.6 & 89.8 & 64.9 & 72.4 & \textbf{86.3} & 25.3 & 68.1 & 45.7 & 51.5 & 69.4 & 66.80 & \textbf{65.4} & 49.4 & 49.8 & 77.0 & 60.39 \\
     {  + GPT score mean \citep{menon2022visual}} &      5.8 &  71.7 & 94.3 & 89.9 & 64.5 & 72.1 & 86.0 & 24.5 & \textbf{68.6} & \textbf{46.6} & 53.8 & 68.4 & 66.86 & 64.9 & 49.4 & 48.8 & 76.6 & 59.92 \\
     {  + Random descriptors} &      32 &  71.6 & 94.6 & 89.3 & 64.7 & 72.1 & 86.0 & 25.3 & 67.5 & 45.4 & 55.2 & 68.8 & 66.89 & 64.8 & 49.9 & 50.2 & 77.9 & 60.69 \\
     {  + Waffle CLIP \citep{roth2023waffling}} &      32 &  71.6 & 94.1 & 89.8 & 65.0 & 72.6 & 86.1 & 26.1 & 67.7 & 45.0 & 50.9 & 68.4 & 66.58 & 65.1 & 49.7 & 50.3 & 77.4 & 60.65 \\
     {  + Descriptor soup (ours)} &      16.7 &  72.1 & \textbf{94.7} & 89.9 & 65.0 & 72.4 & \textbf{86.3} & 25.6 & 68.0 & 45.6 & 53.9 & \textbf{69.5} & 67.10 & 65.3 & 49.7 & 50.1 & 77.7 & 60.70 \\
     {   $\quad$  + offset trick (ours)} &      100 &  72.1 & 94.1 & \textbf{90.4} & \textbf{66.3} & \textbf{73.3} & \textbf{86.3} & \textbf{26.1} & 67.8 & 46.4 & 55.0 & 69.4 & \textbf{67.51} & 65.3 & 49.8 & 50.8 & 78.2 & 61.01 \\
     {  + Word soup centroids (ours)} &      8 &  71.8 & 94.4 & \textbf{90.4} & 65.0 & 72.3 & 86.1 & 25.3 & 68.2 & 45.5 & 55.4 & 69.1 & 67.16 & 65.2 & 50.2 & 50.7 & \textbf{78.7} & 61.22 \\
     {  + Word soup score mean (ours)} &      8 &  71.7 & 94.5 & 90.2 & 65.1 & 72.4 & 86.2 & 25.6 & 68.1 & 45.6 & \textbf{57.3} & 69.3 & 67.43 & 65.3 & \textbf{50.3} & \textbf{50.9} & \textbf{78.7} & \textbf{61.32} \\

     {  + \textcolor{gray}{Descriptor soup upper bound}} & 11 & \textcolor{gray}{71.7} & \textcolor{gray}{94.4} & \textcolor{gray}{90.2} & \textcolor{gray}{66.5} & \textcolor{gray}{72.9} & \textcolor{gray}{86.1} & \textcolor{gray}{26.3} & \textcolor{gray}{67.4} & \textcolor{gray}{46.4} & \textcolor{gray}{57.2} & \textcolor{gray}{68.6} & \textcolor{gray}{67.62} & \textcolor{gray}{64.9} & \textcolor{gray}{49.7} & \textcolor{gray}{50.9} & \textcolor{gray}{78.6} & \textcolor{gray}{61.01} \\

     \midrule

     ProGrad \cite{zhu2023prompt} &      1 &  69.8 & 94.4 & \textbf{91.5} & \textbf{65.8} & \textbf{72.4} & \textbf{86.4} & \textbf{25.3} & 66.6 & 47.2 & 46.3 & 69.0 & 66.48 & 63.2 & 48.2 & 48.6 & 75.9 & 58.96 \\
     KgCoOp \cite{kan2023knowledge} &      1 &  69.2 & 94.3 & 89.9 & 63.9 & 71.0 & 85.7 & 23.7 & 66.2 & 44.4 & 54.4 & 68.3 & 66.16 & 62.3 & 48.0 & 48.8 & 75.5 & 58.64 \\
     ProDA \cite{lu2022prompt} &      32 &  70.0 & 94.2 & 90.2 & 64.7 & 70.8 & 85.7 & 23.1 & 67.0 & 45.8 & 51.4 & \textbf{69.4} & 66.23 & 63.0 & 48.1 & 48.4 & 75.7 & 58.83 \\
     Vanilla CoOp \citep{zhou2022coop} &      1 &  70.0 & 94.6 & 91.2 & 65.4 & 71.2 & 86.3 & 24.6 & 66.9 & \textbf{48.0} & 48.3 & 68.7 & 66.52 & 63.2 & 48.4 & 49.2 & 76.2 & 59.25 \\
     {  + Word soup score mean (ours)} &      8 &  \textbf{70.2} & \textbf{94.7} & 90.9 & 65.4 & 72.0 & 86.0 & 25.0 & \textbf{67.7} & 45.9 & \textbf{56.2} & 69.2 & \textbf{67.30} & \textbf{63.6} & \textbf{49.3} & \textbf{50.1} & \textbf{77.9} & \textbf{60.25}  \\

    

     \midrule

     Vanilla MaPLe \citep{khattak2023maple} &      1 &  70.7 & 93.7 & \textbf{91.2} & \textbf{65.4} & \textbf{71.9} & \textbf{86.2} & \textbf{25.0} & \textbf{67.2} & \textbf{46.2} & 48.6 & \textbf{68.9} & 66.44 & 63.9 & 48.6 & 48.4 & 76.3 & 59.32 \\
     {  + Word soup score mean (ours)} &      8 &  \textbf{70.8} & \textbf{94.1} & 91.2 & 65.2 & 71.8 & 85.8 & 24.0 & 67.0 & 46.0 & \textbf{53.5} & 68.0 & \textbf{66.65} & \textbf{64.0} & \textbf{49.6} & \textbf{49.2} & \textbf{77.9} & \textbf{60.20} \\


     \midrule
    
     Vanilla CLIPood \citep{shu2023clipood} &      1 &  \textbf{72.9} & \textbf{94.8} & 89.8 & \textbf{64.9} & 72.2 & 85.9 & \textbf{25.8} & 67.8 & \textbf{46.4} & 48.7 & 68.7 & 66.50 & \textbf{66.0} & 49.5 & 49.5 & 76.9 & 60.47 \\
     {  + Word soup score mean (ours)} &      8 &  72.0 & 94.4 & \textbf{90.8} & 64.8 & \textbf{72.4} & \textbf{86.0} & 25.4 & \textbf{67.9} & 46.0 & \textbf{57.6} & \textbf{68.9} & \textbf{67.42} & 65.5 & \textbf{50.2} & \textbf{50.8} & \textbf{78.5} & \textbf{61.23} \\
     
\bottomrule
  \end{tabular}
  \caption{Comparison with few-shot methods and few-shot methods stacked with ZS methods. $\dagger$ indicates author-reported numbers on the same datasets with the same train-test splits. Other numbers are from our reproductions using our github code. We tune all baselines on a withheld validation set, so our numbers are different from published numbers. The descriptor soup upper bound was trained to maximize average cross-dataset accuracy (on test data); this loosely approximates the maximally achievable accuracy on these benchmarks without using extra information. All other methods were \emph{trained on 3 random 16-shot splits of ImageNet}. $m$ indicates number of descriptors used. All methods are evaluated on top of 3 models finetuned with different random seeds. Due to space limitations, we only compare with ZS baselines stacked on top of the CE-finetuned few-shot model, since this is the best finetuned model. Either our descriptor soup with the offset trick or our word soup achieves the best accuracy on most datasets. Finally, we stack our word soup method on top of CoOp, MaPLe, and CLIPood finetuned models to show that word soup is complementary to most existing robust finetuning methods. Average of 3 trials. This is an expanded version of Table \ref{tab:ft} in the main paper. }
  \label{tab:ft-appendix}
\end{table*}
\setlength\tabcolsep{6 pt}

\setlength\tabcolsep{4 pt}
\begin{table*}[t!]
\scriptsize
\centering
\begin{tabular}{l c ccc ccc ccccc >{\columncolor[gray]{0.9}}ccccc >{\columncolor[gray]{0.9}}c}
\toprule
&& \textbf{Source} & \multicolumn{11}{c}{\textbf{Cross-dataset Evaluation Targets}} & \multicolumn{5}{c}{\textbf{Domain Generalization Targets}} \\

\cmidrule(lr){3-3} \cmidrule(lr){4-14} \cmidrule(lr){15-19}

     & \rotatebox{90}{\parbox{1cm}{parameters \\ (thousands)}} & \rotatebox{90}{ INet } & \rotatebox{90}{ Caltech } & \rotatebox{90}{ Pets } & \rotatebox{90}{ Cars } & \rotatebox{90}{ Flowers } & \rotatebox{90}{ Food } & \rotatebox{90}{ Aircraft } & \rotatebox{90}{ SUN } & \rotatebox{90}{ DTD } & \rotatebox{90}{ EuroSAT } & \rotatebox{90}{ UCF } & \rotatebox{90}{ \textbf{Average} } & \rotatebox{90}{ INet-V2 } & \rotatebox{90}{ INet-Sketch } & \rotatebox{90}{ INet-A } & \rotatebox{90}{ INet-R } & \rotatebox{90}{ \textbf{Average} }\\
     
    \midrule
    
     VPT shallow 1 token &      0.768 &  68.7 & 93.8 & 90.0 & 65.1 & 69.5 & 85.3 & 24.2 & 66.0 & 44.7 & 41.9 & 67.8 & 64.84 & 62.1 & 47.9 & 47.9 & 76.7 & 58.67 \\
     VPT shallow 2 tokens &      2 &  68.7 & 93.8 & 90.0 & 65.2 & 69.5 & 85.2 & 24.2 & 66.2 & 44.8 & 42.3 & 67.1 & 64.84 & 62.2 & 48.0 & 47.3 & 76.7 & 58.54 \\
     VPT shallow 3 tokens &      2 &  68.7 & 93.9 & 90.0 & 65.6 & 70.2 & 85.3 & 24.8 & 66.2 & 44.7 & 43.8 & 67.5 & 65.20 & 62.4 & 48.1 & 47.0 & 76.6 & 58.52 \\
     VPT shallow 3 tokens &      2 &  68.6 & 93.8 & 89.5 & 64.8 & 70.1 & 85.3 & 24.1 & 66.1 & 44.5 & 45.4 & 67.7 & 65.12 & 62.1 & 48.0 & 47.1 & 76.4 & 58.41 \\
     VPT deep 2 layers &      5 &  68.8 & 93.5 & 89.7 & 65.0 & 70.3 & 85.4 & 24.0 & 65.9 & 44.7 & 49.3 & 67.6 & 65.54 & 62.2 & 48.2 & 46.9 & 76.6 & 58.47 \\
     VPT deep 3 layers &      7 &  68.7 & 93.5 & 89.4 & 65.3 & 70.4 & 85.3 & 24.2 & 66.2 & 44.8 & 45.0 & 67.5 & 65.16 & 62.3 & 48.2 & 46.8 & 76.4 & 58.42 \\
     \midrule
     MaPLe 1 layer &      396 &  70.1 & 94.2 & 91.1 & 64.3 & 71.1 & 86.1 & 24.5 & 67.0 & 47.3 & 51.8 & 68.6 & 66.61 & 63.4 & 48.4 & 48.8 & 76.3 & 59.22 \\
     MaPLe 2 layers &      397 &  70.4 & 93.6 & \textbf{91.8} & 64.3 & 71.3 & 85.9 & 24.7 & 67.0 & 46.9 & 48.1 & 68.5 & 66.21 & 63.7 & 48.3 & 49.2 & 76.1 & 59.34 \\
     MaPLe 3 layers &      399 &  \textbf{70.7} & 93.7 & 91.2 & 65.4 & 71.9 & 86.2 & 25.0 & 67.2 & 46.2 & 48.6 & 68.9 & 66.44 & \textbf{63.9} & 48.6 & 48.4 & 76.3 & 59.32 \\
     \midrule
     bitfit last layer &      17 &  68.3 & 94.1 & 89.5 & 65.2 & 71.4 & 85.9 & 24.9 & 65.7 & 44.7 & 46.9 & 67.9 & 65.62 & 61.7 & 48.0 & 48.5 & 75.9 & 58.51 \\
     bitfit last 2 layers &      34 &  68.8 & 93.9 & 89.9 & 65.3 & 71.4 & 85.9 & 25.1 & 66.4 & 45.1 & 47.4 & 68.4 & 65.88 & 62.1 & 48.6 & 48.5 & 76.6 & 58.93 \\
     bitfit last 3 layers &      51 &  69.1 & 93.9 & 90.0 & 65.3 & 71.7 & 85.8 & 25.0 & 66.7 & 45.4 & 48.3 & 68.4 & 66.05 & 62.6 & 48.7 & 48.5 & 76.8 & 59.12 \\
     \midrule
     CoOp 1 token &      0.512 &  69.4 & 94.3 & 91.4 & 64.4 & 71.7 & 86.3 & 24.6 & 67.2 & 47.3 & 49.1 & 68.5 & 66.49 & 63.1 & 48.2 & 49.0 & 76.1 & 59.08 \\
     CoOp 2 tokens &      1 &  69.9 & \textbf{94.6} & 91.6 & 65.5 & 72.0 & 86.1 & 25.0 & 66.8 & \textbf{48.2} & 49.6 & \textbf{69.4} & 66.89 & 63.2 & 48.5 & 48.8 & 76.3 & 59.20 \\
     CoOp 3 tokens &      2 &  70.2 & 94.5 & 91.0 & \textbf{66.0} & 71.6 & 86.3 & 24.6 & 66.8 & 47.6 & 49.0 & 68.9 & 66.63 & 63.4 & 48.5 & 49.5 & 76.3 & 59.45 \\
     \midrule
     ProGrad 1 token &      0.512 &  69.4 & 94.2 & 91.0 & 65.6 & \textbf{72.7} & 86.4 & 25.1 & 66.2 & 46.0 & 48.2 & 68.5 & 66.39 & 62.8 & 48.1 & 48.5 & 75.7 & 58.77 \\
     ProGrad 2 tokens &      1 &  69.5 & 94.1 & 90.8 & 65.7 & 72.6 & 86.3 & 24.8 & 66.5 & 45.5 & 47.7 & 68.7 & 66.28 & 62.8 & 48.0 & 48.5 & 75.7 & 58.75 \\
     ProGrad 3 tokens &      2 &  69.8 & 94.4 & 91.5 & 65.8 & 72.4 & 86.4 & 25.3 & 66.6 & 47.2 & 46.3 & 69.0 & 66.48 & 63.2 & 48.2 & 48.6 & 75.9 & 58.96 \\
     \midrule
     KgCoOp 1 token &      0.512 &  68.6 & 93.4 & 89.4 & 63.4 & 70.9 & 85.9 & 23.8 & 65.6 & 44.9 & 52.5 & 68.1 & 65.80 & 62.0 & 47.8 & 49.1 & 75.7 & 58.63 \\
     KgCoOp 2 tokens &      1 &  69.0 & 93.3 & 89.3 & 62.8 & 70.2 & 85.8 & 23.8 & 66.0 & 45.4 & 53.0 & 69.0 & 65.85 & 62.4 & 48.0 & 49.1 & 75.9 & 58.85 \\
     KgCoOp 3 tokens &      2 &  69.2 & 94.3 & 89.9 & 63.9 & 71.0 & 85.7 & 23.7 & 66.2 & 44.4 & 54.4 & 68.3 & 66.16 & 62.3 & 48.0 & 48.8 & 75.5 & 58.64 \\
     \midrule
     ProDA ensemble size 4 &      20 &  70.5 & 94.3 & 90.4 & 65.3 & 71.2 & 86.1 & 24.9 & 67.2 & 46.4 & 50.4 & 69.4 & 66.54 & 63.6 & 48.6 & 49.4 & 76.0 & 59.43 \\
     ProDA ensemble size 8 &      41 &  70.1 & 93.8 & 90.3 & 65.1 & 71.0 & 85.8 & 24.9 & 67.4 & 45.5 & 49.4 & 68.4 & 66.15 & 63.3 & 48.8 & 49.5 & 76.6 & 59.55 \\
     ProDA ensemble size 16 &      82 &  69.9 & 94.3 & 90.5 & 64.5 & 70.8 & 85.6 & 24.3 & 66.6 & 45.2 & 48.4 & 68.8 & 65.90 & 63.1 & 48.4 & 48.9 & 76.1 & 59.13 \\
     ProDA ensemble size 32 &      164 &  70.0 & 94.2 & 90.2 & 64.7 & 70.8 & 85.7 & 23.1 & 67.0 & 45.8 & 51.4 & 69.4 & 66.23 & 63.0 & 48.1 & 48.4 & 75.7 & 58.83 \\
     ProDA ensemble size 64 &      328 &  69.4 & 94.4 & 90.0 & 64.5 & 69.5 & 85.1 & 22.7 & 66.4 & 44.9 & 49.6 & 67.8 & 65.49 & 62.7 & 48.0 & 48.7 & 76.2 & 58.91 \\
     \midrule
     CLIP-adapter reduction=128 &      4 &  67.1 & 93.3 & 89.0 & 65.3 & 70.9 & 85.7 & 25.1 & 63.3 & 43.5 & 46.6 & 67.4 & 65.00 & 60.9 & 46.6 & 47.2 & 74.1 & 57.18 \\
     CLIP-adapter reduction=64 &      8 &  67.1 & 93.3 & 88.8 & 65.4 & 71.1 & 85.7 & 24.9 & 63.3 & 43.5 & 46.5 & 67.2 & 64.97 & 60.9 & 46.5 & 47.2 & 74.0 & 57.17 \\
     CLIP-adapter reduction=32 &      16 &  67.4 & 93.2 & 88.4 & 65.2 & 70.1 & 85.6 & 24.9 & 64.1 & 44.0 & 46.3 & 66.8 & 64.84 & 60.9 & 46.9 & 47.9 & 74.5 & 57.55 \\
     CLIP-adapter reduction=16 &      33 &  67.6 & 93.3 & 88.3 & 64.9 & 70.1 & 85.6 & 24.5 & 64.4 & 43.9 & 46.7 & 66.8 & 64.86 & 61.2 & 47.2 & 48.4 & 75.1 & 57.98 \\
     CLIP-adapter reduction=8 &      66 &  67.9 & 93.4 & 88.7 & 65.4 & 70.2 & 85.7 & 24.8 & 65.1 & 44.3 & 46.6 & 66.7 & 65.09 & 61.5 & 47.5 & 48.5 & 75.3 & 58.21 \\
     CLIP-adapter reduction=4 &      131 &  67.8 & 93.4 & 89.0 & 65.2 & 70.2 & 85.7 & 24.5 & 65.2 & 44.2 & 46.0 & 66.8 & 65.02 & 61.5 & 47.5 & 48.3 & 75.1 & 58.12 \\
     \midrule
     SSF last layer &      12 &  68.1 & 94.0 & 89.5 & 65.4 & 71.0 & 85.7 & 24.7 & 65.6 & 45.3 & 51.6 & 68.5 & 66.13 & 61.6 & 47.8 & 46.4 & 75.7 & 57.87 \\
     SSF last 2 layers &      25 &  68.5 & 94.1 & 89.9 & 65.1 & 71.2 & 85.8 & 24.8 & 66.3 & 45.9 & 49.1 & 68.2 & 66.04 & 62.1 & 48.3 & 47.2 & 76.3 & 58.46 \\
     SSF last 3 layers &      37 &  68.5 & 94.2 & 89.5 & 64.9 & 71.2 & 85.3 & 24.4 & 66.2 & 45.8 & 49.3 & 67.8 & 65.86 & 62.1 & 48.1 & 47.2 & 76.3 & 58.44 \\
     \midrule
     LoRA rank=1 &      18 &  67.3 & 93.5 & 89.3 & 65.4 & 71.3 & 85.7 & 25.1 & 64.2 & 44.4 & 47.9 & 67.6 & 65.43 & 61.4 & 47.1 & 46.9 & 74.9 & 57.59 \\
     LoRA rank=2 &      37 &  67.6 & 93.7 & 90.0 & 65.7 & 71.2 & 85.7 & 25.3 & 65.6 & 45.9 & 49.6 & 67.8 & 66.05 & 61.9 & 47.7 & 45.3 & 75.6 & 57.62 \\
     LoRA rank=4 &      74 &  67.6 & 93.8 & 90.1 & 65.7 & 71.5 & 85.7 & 25.2 & 65.4 & 46.0 & 50.9 & 67.7 & 66.19 & 61.8 & 47.7 & 46.2 & 76.0 & 57.93 \\
     LoRA rank=8 &      147 &  68.0 & 93.9 & 90.0 & 65.7 & 71.4 & 85.4 & 25.5 & 65.9 & 46.3 & 52.6 & 67.2 & 66.39 & 61.9 & 47.1 & 42.2 & 74.4 & 56.40 \\
     \midrule
     ResBlock-adapter reduction=128 &      55 &  68.0 & 93.8 & 89.2 & 64.0 & 71.1 & 84.7 & 23.3 & 65.1 & 45.3 & 46.0 & 67.6 & 65.01 & 61.2 & 47.4 & 47.2 & 75.5 & 57.81 \\
     ResBlock-adapter reduction=64 &      111 &  68.8 & 94.0 & 89.7 & 64.2 & 70.8 & 85.0 & 23.5 & 65.8 & 45.5 & 46.9 & 68.0 & 65.35 & 61.8 & 48.0 & 48.0 & 76.3 & 58.52 \\
     ResBlock-adapter reduction=32 &      221 &  69.1 & 94.2 & 90.0 & 64.4 & 71.4 & 85.3 & 23.2 & 66.1 & 45.2 & 46.8 & 67.4 & 65.41 & 62.5 & 48.1 & 48.3 & 76.8 & 58.94 \\
     ResBlock-adapter reduction=16 &      442 &  69.3 & 94.2 & 89.9 & 64.2 & 71.3 & 85.3 & 23.8 & 66.4 & 45.6 & 47.5 & 67.9 & 65.60 & 62.8 & 48.4 & 48.4 & 76.9 & 59.12 \\
     ResBlock-adapter reduction=8 &      885 &  69.5 & 94.1 & 89.5 & 64.6 & 71.3 & 85.6 & 23.6 & 66.6 & 44.8 & 45.3 & 67.9 & 65.33 & 63.0 & 48.6 & 48.8 & 77.0 & 59.36 \\
     ResBlock-adapter reduction=4 &      1769 &  69.7 & 94.1 & 89.5 & 64.8 & 71.2 & 85.5 & 24.0 & 66.8 & 44.9 & 46.8 & 67.8 & 65.55 & 63.1 & 48.7 & 49.0 & 77.1 & 59.48 \\
     \midrule
     Word Soup $m=1$ &      \textbf{0.012} &  68.6 & 93.9 & 89.2 & 64.6 & 71.8 & 86.0 & 24.7 & 65.9 & 44.2 & 48.0 & 67.7 & 65.61 & 62.1 & 47.9 & 49.7 & 76.3 & 59.01 \\
     Word Soup $m=2$ &      0.024 &  69.0 & 94.1 & 90.3 & 65.6 & 72.5 & 86.0 & 25.5 & 66.9 & 45.0 & 52.0 & 68.6 & 66.64 & 62.4 & 48.8 & 50.2 & 76.6 & 59.50 \\
     Word Soup $m=4$ &      0.048 &  69.3 & 94.1 & 89.9 & 65.9 & 72.4 & \textbf{86.5} & 25.7 & 67.1 & 45.8 & 53.6 & 68.7 & 66.96 & 62.9 & 48.9 & 50.3 & 77.2 & 59.80 \\
     Word Soup $m=8$ &      0.096 &  69.4 & 94.1 & 89.9 & 65.7 & 72.5 & 86.4 & 25.9 & 67.0 & 44.9 & 54.6 & 68.8 & 66.99 & 63.1 & 49.0 & 50.5 & 77.3 & 59.95 \\
     Word Soup $m=16$ &      0.192 &  69.5 & 94.0 & 89.9 & 65.9 & 72.5 & 86.3 & 26.1 & 67.4 & 45.2 & 54.8 & 68.8 & 67.08 & 63.2 & 49.0 & 50.7 & 77.2 & 60.02 \\
     Word Soup $m=32$ &      0.384 &  69.6 & 94.2 & 89.9 & 65.9 & 72.4 & 86.5 & \textbf{26.2} & 67.4 & 45.1 & 54.7 & 69.0 & 67.12 & 63.2 & 49.0 & 50.6 & 77.3 & 60.04 \\
     Word Soup $m=64$ &      0.767 &  69.5 & 94.1 & 90.0 & 65.9 & 72.5 & 86.4 & 26.2 & 67.4 & 45.2 & 55.1 & 69.0 & 67.17 & 63.3 & 49.1 & \textbf{50.7} & 77.4 & 60.11 \\
     \midrule
     Word Soup + CoOp $m=4$ &      2 &  70.2 & 94.5 & 91.0 & 65.6 & 72.3 & 86.0 & 25.1 & 67.7 & 45.7 & \textbf{56.1} & 68.6 & 67.26 & 63.7 & 49.3 & 50.1 & 77.9 & 60.26 \\
     Word Soup + CoOp $m=8$ &      2 &  70.2 & 94.4 & 91.0 & 65.3 & 72.1 & 86.1 & 25.2 & 67.7 & 45.5 & 55.5 & 68.7 & 67.15 & 63.5 & 49.3 & 50.2 & \textbf{78.0} & 60.25 \\
     Word Soup + CoOp $m=16$ &      2 &  70.2 & 94.5 & 91.0 & 65.7 & 72.6 & 86.1 & 24.9 & \textbf{67.8} & 45.6 & 55.5 & 69.2 & \textbf{67.30} & 63.7 & \textbf{49.5} & 50.5 & 77.9 & \textbf{60.39} \\
\bottomrule
  \end{tabular}
  \caption{Detailed numerical results for PEFT comparison in Fig. \ref{fig:parameter_efficiency}. Average of 3 trials. These results are plotted in Figure \ref{fig:parameter_efficiency} of the main paper. Also reference Section 7 (Results) for a discussion. }
  \label{tab:parameter_efficiency_detailed}
\end{table*}
\setlength\tabcolsep{6 pt}

\setlength\tabcolsep{5.3 pt}
\begin{table*}[t!]
\scriptsize
\centering
\begin{tabular}{l c ccc ccc ccccc >{\columncolor[gray]{0.9}}ccccc >{\columncolor[gray]{0.9}}c}
\toprule
&& \textbf{Source} & \multicolumn{11}{c}{\textbf{Cross-dataset Evaluation Targets}} & \multicolumn{5}{c}{\textbf{Domain Generalization Targets}} \\

\cmidrule(lr){3-3} \cmidrule(lr){4-14} \cmidrule(lr){15-19}

     & $m$ & \rotatebox{90}{ INet } & \rotatebox{90}{ Caltech } & \rotatebox{90}{ Pets } & \rotatebox{90}{ Cars } & \rotatebox{90}{ Flowers } & \rotatebox{90}{ Food } & \rotatebox{90}{ Aircraft } & \rotatebox{90}{ SUN } & \rotatebox{90}{ DTD } & \rotatebox{90}{ EuroSAT } & \rotatebox{90}{ UCF } & \rotatebox{90}{ \textbf{Mean} } & \rotatebox{90}{ INet-V2 } & \rotatebox{90}{ Sketch } & \rotatebox{90}{ INet-A } & \rotatebox{90}{ INet-R } & \rotatebox{90}{ \textbf{Mean} }\\
     
    \midrule
    \multicolumn{19}{c}{\textbf{Open-AI CLIP ViT-B/32}} \\
    \midrule
    
    ZS &      1 &  61.9 & 91.5 & 87.4 & 60.3 & 66.4 & 80.2 & 19.1 & 62.2 & 42.3 & 40.3 & 63.5 & 61.32 & 54.6 & 40.7 & 29.1 & 66.3 & 47.68 \\
     GPT score mean &      5.8 &  63.0 & 91.8 & \textbf{88.1} & 60.0 & 66.6 & 80.2 & 19.1 & 64.4 & 43.1 & 36.2 & 62.7 & 61.22 & 55.4 & 41.0 & 29.4 & 65.9 & 47.95 \\
     Waffle CLIP  &      16 &  63.3 & \textbf{91.8} & 88.0 & \textbf{60.9} & \textbf{67.4} & 80.4 & 19.6 & 63.8 & 41.7 & 44.8 & 63.0 & 62.13 & 55.8 & 41.6 & 31.1 & 67.8 & 49.07 \\
     Desc. soup + offsets &      100 &  64.1 & 91.5 & 87.7 & 60.7 & 66.9 & 80.4 & \textbf{19.9} & 64.4 & \textbf{43.6} & \textbf{48.3} & \textbf{64.5} & \textbf{62.79} & 56.5 & \textbf{42.6} & 31.8 & \textbf{69.3} & \textbf{50.05} \\
     Word soup &      8 &  \textbf{64.5} & 91.5 & 88.0 & 60.4 & 67.0 & \textbf{80.9} & 19.3 & \textbf{64.6} & 42.0 & 45.5 & 63.2 & 62.24 & \textbf{56.9} & 42.5 & \textbf{32.0} & 68.7 & 50.00 \\

     \midrule
    \multicolumn{19}{c}{\textbf{Open CLIP ViT-L/14}} \\
    \midrule
    
     ZS &      1 &  73.3 & 96.4 & \textbf{92.9} & 92.0 & 75.8 & 85.7 & 34.1 & 72.7 & 57.3 & 52.1 & 72.1 & 73.11 & 65.6 & 61.0 & 47.2 & 85.7 & 64.88 \\
     GPT score mean &      5.8 &  73.6 & \textbf{96.7} & 92.8 & 91.2 & \textbf{76.5} & 85.3 & 33.7 & 72.7 & 58.6 & 51.6 & 71.7 & 73.08 & 66.1 & 61.2 & 47.5 & 85.1 & 64.96 \\
     Waffle CLIP  &      16 &  72.7 & 96.1 & 92.4 & 91.7 & 76.4 & 85.8 & 34.4 & 72.4 & 58.6 & 52.2 & 72.5 & 73.25 & 65.3 & 60.7 & 46.5 & 85.4 & 64.47 \\
     Desc. soup + offsets &      100 &  74.0 & 96.6 & 92.8 & 92.0 & 76.3 & 85.5 & 34.5 & 72.7 & \textbf{59.1} & 50.0 & 72.3 & 73.19 & 66.0 & \textbf{61.9} & \textbf{48.7} & \textbf{86.6} & \textbf{65.81} \\
     Word soup &      8 &  \textbf{74.3} & 96.5 & 92.1 & \textbf{92.2} & 76.0 & \textbf{86.0} & \textbf{35.0} & \textbf{73.6} & 58.5 & \textbf{52.9} & \textbf{73.0} & \textbf{73.56} & \textbf{66.8} & 61.6 & 48.2 & 86.3 & 65.73 \\

     \midrule
    \multicolumn{19}{c}{\textbf{Open CLIP CoCa-L/14}} \\
    \midrule
    
     ZS &      1 &  75.1 & \textbf{97.6} & 93.8 & 92.7 & 77.3 & 87.5 & 36.6 & 73.6 & 57.2 & 58.5 & 73.4 & 74.82 & 67.5 & 63.5 & 53.8 & 87.0 & 67.94 \\
     GPT score mean &      5.8 &  74.9 & \textbf{97.6} & 93.7 & 92.4 & 76.2 & 87.3 & 36.3 & 73.9 & 58.9 & \textbf{64.9} & 73.6 & 75.48 & 67.6 & 63.5 & 52.8 & 86.8 & 67.67 \\
     Waffle CLIP  &      16 &  75.0 & 97.5 & 93.9 & 92.7 & 77.3 & 87.5 & 37.4 & 73.1 & 57.5 & 63.0 & 73.9 & 75.37 & 67.5 & 63.8 & 52.8 & 87.3 & 67.85 \\
     Desc. soup + offsets &      100 &  75.5 & 97.5 & \textbf{93.9} & 92.6 & 77.5 & 87.3 & 37.2 & 73.8 & \textbf{61.1} & 63.6 & \textbf{75.0} & 75.95 & 68.0 & \textbf{64.2} & 53.2 & 87.9 & 68.32 \\
     Word soup &      8 &  \textbf{75.9} & 97.5 & 93.8 & \textbf{92.8} & \textbf{77.8} & \textbf{87.7} & \textbf{38.4} & \textbf{74.1} & 60.5 & 63.5 & 74.7 & \textbf{76.08} & \textbf{68.8} & 64.0 & \textbf{54.3} & \textbf{87.9} & \textbf{68.73} \\

     \midrule
    \multicolumn{19}{c}{\textbf{Open CLIP ViT-g/14}} \\
    \midrule
    
     ZS &      1 &  77.7 & 97.7 & 93.6 & 93.5 & \textbf{81.6} & \textbf{90.0} & 44.1 & 74.3 & 65.3 & 55.8 & \textbf{80.0} & 77.58 & 70.4 & 66.4 & 59.7 & 89.0 & 71.37 \\
     GPT score mean &      5.8 &  77.6 & 97.2 & 93.7 & 93.6 & 81.4 & 89.6 & 43.1 & 74.7 & 63.1 & 58.7 & 76.3 & 77.14 & 71.0 & 66.3 & 58.8 & 88.9 & 71.26 \\
     Waffle CLIP  &      16 &  77.3 & 97.8 & 93.5 & 93.7 & 81.3 & 89.8 & \textbf{44.1} & 74.1 & 65.8 & 58.0 & 78.9 & 77.72 & 70.1 & 65.9 & 59.0 & 88.9 & 70.99 \\
     Desc. soup + offsets &      100 &  78.0 & \textbf{97.8} & \textbf{94.1} & \textbf{93.9} & 80.7 & 89.2 & 43.1 & \textbf{75.0} & \textbf{67.0} & \textbf{60.4} & 79.2 & 78.04 & 71.5 & \textbf{67.2} & \textbf{60.2} & \textbf{90.0} & \textbf{72.21} \\
     Word soup &      8 &  \textbf{78.4} & 97.6 & 93.7 & 93.9 & 81.4 & 89.8 & 44.0 & 75.0 & 66.0 & 60.0 & 79.5 & \textbf{78.09} & \textbf{71.6} & 67.1 & 60.0 & 89.6 & 72.05 \\
     
\bottomrule
  \end{tabular}
  \caption{ Detailed numerical results for different model scales. This is an expanded version of Table \ref{tab:scaling}. Average of 3 trials. }
  \label{tab:model-scaling-detailed}
\end{table*}
\setlength\tabcolsep{6 pt}


\end{document}